\documentclass[letterpaper]{article} 
\usepackage{aaai2026}  

\usepackage{verbatim}
\usepackage{microtype}
\usepackage{array}
\usepackage{tabularx}
\usepackage{booktabs}
\usepackage{xcolor}
\usepackage{makecell}
\usepackage{adjustbox}
\usepackage{longtable}
\usepackage{multirow}
\usepackage{lineno}
\usepackage{subcaption}
\usepackage{amsmath}
\usepackage{cleveref}
\usepackage{algorithm}
\usepackage{algorithmic}
\usepackage[ruled,vlined,algo2e]{algorithm2e}
\usepackage{todonotes}
\usepackage{inconsolata}
\usepackage{paralist}
\usepackage{amssymb}
\usepackage{pifont}

\newcommand\datasetname{\textsc{PragWorld}}
\definecolor{darkgreen}{rgb}{0,0.5,0}
\definecolor{darkblue}{rgb}{0, 0, 0.5}
\newcommand{\cmark}{\textcolor{darkgreen}{\ding{51}}}
\newcommand{\xmark}{\textcolor{red}{\ding{55}}}

\usepackage{adjustbox}
\usepackage{amssymb}

\usepackage{times}  
\usepackage{helvet}  
\usepackage{courier}  
\usepackage[hyphens]{url}  
\usepackage{graphicx} 
\usepackage{natbib}  
\usepackage{caption} 
\usepackage{algorithm}
\usepackage{algorithmic}

\usepackage{newfloat}
\usepackage{listings}
\DeclareCaptionStyle{ruled}{labelfont=normalfont,labelsep=colon,strut=off} 
\floatstyle{ruled}
\newfloat{listing}{tb}{lst}{}
\floatname{listing}{Listing}
\title{\textsc{PragWorld}:  A Benchmark Evaluating LLMs' \textit{Local World Model} under Minimal Linguistic Alterations and Conversational Dynamics}

\author{
  Sachin Vashistha$^1$,  
  Aryan Bibhuti$^1$,  
  Atharva Naik$^{2}$,  
  Martin Tutek$^3$,
  Somak Aditya$^1$ 
  \\
  }
  \affiliations{
  $^1$Indian Institute of Technology Kharagpur, India\\ 
  $^2$ LTI, Carnegie Mellon University, $^3$ University of Zagreb
 }
\usepackage{bibentry}

\begin{document}

\maketitle

\begin{abstract}
Real-world conversations are rich with pragmatic elements, such as entity mentions, references, and implicatures.
Understanding such nuances is a requirement for successful natural communication, and often requires building a local \textit{world model} which encodes such elements and captures the dynamics of their evolving states.
However, it is not well-understood whether language models (LMs) construct or maintain a robust implicit representation of conversations.
In this work, we evaluate the ability of LMs to encode and update their internal world model in dyadic conversations and test their \textit{malleability} under linguistic alterations.
To facilitate this, we apply seven minimal linguistic alterations to conversations sourced from popular conversational QA datasets and construct a benchmark with two variants (i.e., Manual and Synthetic) comprising yes-no questions. 
We evaluate nine open and one closed source LMs and observe that they struggle to maintain robust accuracy. 
Our analysis unveils that LMs struggle to memorize crucial details, such as tracking entities under linguistic alterations to conversations.
We then propose a dual-perspective interpretability framework 
which identifies transformer layers that are \textit{useful} or \textit{harmful} and highlights linguistic alterations most influenced by harmful layers, typically due to encoding spurious signals or relying on shortcuts.
Inspired by these insights, we propose two layer-regularization based fine-tuning strategies that suppress the effect of the harmful layers.
\end{abstract}

\begin{links}
    \link{Code}{https://github.com/SachinVashisth/PRAGWORLD}
\end{links}

\section{Introduction}
The human ability to comprehend natural language is often owed to our innate skills to utilize relevant world knowledge, ability to map words (symbols) to meaningful concepts (abstract or concrete entities) in the world -- thus creating a mental state of the environment. Forming such a mental model of the world further endows us to suitably calibrate our responses, or actions. On the other hand, Transformer-based language models (LMs) achieve impressive language comprehension ability solely by learning from vast amounts of unstructured text. Thus, NLP researchers face a looming question: \textit{Do LMs construct an implicit world model of the environment described in the input} \cite{bender-koller-2020-climbing}?

\begin{figure}[!t]
\includegraphics[width=\linewidth]{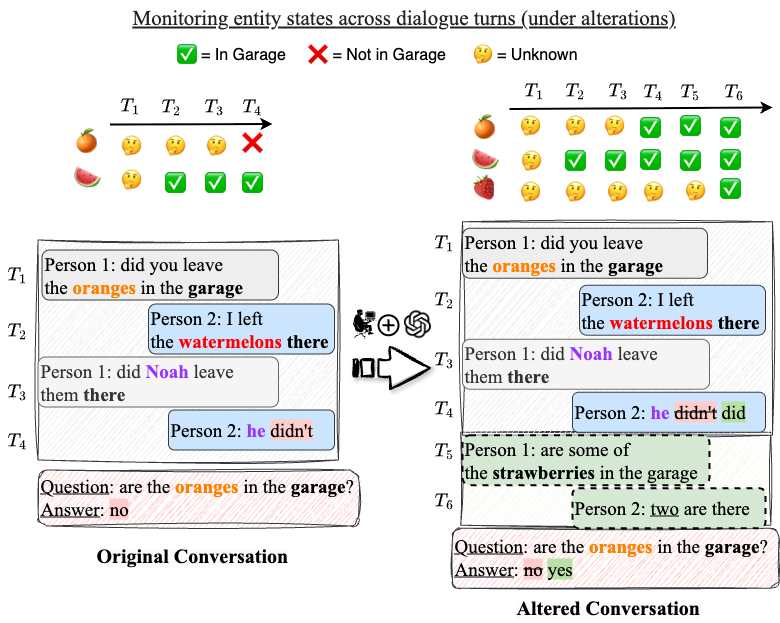}
\caption{Example alteration applied to a conversation from GRICE \citep{zheng-etal-2021-grice}. Linguistic alterations modify objects or their states and introduce or eliminate new agents or entities. \( T_i \;\forall i \in \{1,2,\dots,n\} \text{ denotes the } i^{\mathrm{th}} \text{utterance.} \) 
}
\label{fig:pragworldmotiv}
\end{figure}

Despite being trained only on text, a number of works present evidence that LMs implicitly encode nuanced information about the world such as color \citep{abdou2021can}, gender \citep{bolukbasi2016man} or space \citep{gurnee2023language}, even going so far as representing the board state of chess \citep{toshniwal2022chess,  li2022emergent, kuo2023large, karvonen2024emergent}. 
Such findings suggest that LMs are able to develop latent representations which encode a \textit{world model}, a capability which greatly exceeds the pretraining task of next-token prediction. 
However, \citet{vafa2024evaluating} show that such latent world models of LMs are fragile in tasks such as game playing, logic puzzles and navigation.

Therefore, we proceed with benchmarking the aspect of \textit{malleability} of latent world states -- a notion complementary to fidelity and expressiveness -- by investigating whether an LM is able to adjust accordingly to new information. To model such dynamics, we opt for a \textit{conversational} setting and probe about how knowledge about concrete entities are updated throughout the conversation.
We source seed conversations from existing datasets: \textsc{grice} \citep{zheng-etal-2021-grice} and \textsc{cicero} \citep{ghosal-etal-2022-cicero}. These datasets contain conversations riddled with pragmatic nuances, paired with questions. We introduce various label-preserving or label-altering linguistic alterations to those conversations.
In Figure~\ref{fig:pragworldmotiv}, we show how a simple linguistic alteration can \textit{inadvertently} change the outcome of the question about the entity ``orange''.


Previous benchmarks for entity tracking primarily present the context as a sequence of un-ambiguously stated facts \citep{kim2023entity,tandon-etal-2020-dataset}.
In this work, we take a step further. 
We blend in conversational dynamics, and evaluate entity (or state) tracking ability under the presence of well-defined minimal linguistic alterations: negation, variable swap, quantity change, variable substitution, quantifier change, logical connective change and injecting local knowledge.
We then leverage manual and semi-automatic methods to create alterations of the seed conversations and create $2614$ distinct instances, constituting the manual and synthetic splits of our dataset: \datasetname{}.
We then benchmark a wide range of open and closed-source LMs 
on \datasetname{} and show that the models are not \textit{robustly accurate} --- i.e., accurate on both the original instances and their altered variants -- indicating gaps in reading comprehension or memorization of LMs. 
In order to understand where LMs fail, we design a designed dual-perspective interpretability framework using direct effect patching and MLP zero-out ablation. Using this framework, we trace performance issues to fragility in entity state tracking by identifying harmful model layers. Finally, we design regularization techniques that help reduce the unwanted effect of harmful layers, in turn improving robustness towards proposed linguistic alterations.

Concretely, our \textbf{contributions} are as follows:
\begin{compactitem}
\item We apply seven minimal linguistic alterations to seed conversations from the \textsc{grice} and \textsc{cicero} datasets, creating our benchmark, \textsc{PragWorld}, which evaluates \textit{malleability} of LLMs' internal representations. 
\item  We evaluate a wide range of open and closed source LMs on \textsc{PragWorld} and show they are not \textit{robustly accurate} under linguistic alteration, indicating brittleness of the LM's latent world models.
\item We propose a dual-perspective interpretability framework using \textbf{Direct Effect Patching} and \textbf{MLP zero-out Ablation} to pinpoint layers that encode \textit{useful}, or \textit{harmful} reasoning patterns. 
\item Based on insights of the previous framework, we design and evaluate two regularization techniques: a) \textit{Useful Layer Amplification}, and b) \textit{Harmful Layer Suppression} which help suppress the effect of harmful layers.

\end{compactitem}


\section{Related Work}

\paragraph{Do LMs Encode World Models?}
A number of works explore whether the internal representations of language models recoverably encode something akin to a world model \textit{isomorphic} to the real one \citep{merrill2021provable,patel2022mapping, vafa2024evaluating}.
Works show that attributes such as colors, \citep{abdou2021can}, gender \citep{bolukbasi2016man} and directions \citep{gurnee2023language} can be faithfully recovered from the internal model representations.
In this work, we take these analyses a step further and evaluate whether LMs are able to precisely detect, encode, and update local world states described in the course of dyadic conversations.
In order to keep track of objects, locations, and people in conversations, plan ahead in games, or reason counterfactually, LMs need to create and frequently dynamically update a local world model.
We evaluate the precision of this encoding through minimal linguistic alterations applied to conversations. We compare the complexity evaluated within our benchmark to related works in Supplementary Table \ref{tab:datasetcmparison}.

\paragraph{Complexities of the Conversational Format.}
Conversational dynamics represent the complexity of the real world more accurately compared to static settings by introducing pragmatic elements, implicatures and commonsense inference intertwined with the core task \citep{sap2019atomic,zheng-etal-2021-grice,ghosal-etal-2022-cicero, li2023diplomat}. 
We source our benchmark based on the following two conversational datasets. 
\textsc{grice} \citep{zheng-etal-2021-grice} is an automatically generated dialogue dataset which evaluates the capabilities of LMs for pragmatic reasoning \citep{goodman2016pragmatic}, resolution of implicatures \citep{grice1975logic,borg2009three} and coreferences as well as entity tracking.
\textsc{cicero} \citep{ghosal-etal-2022-cicero} is a conversational dataset necessitating a broad number of inference types and commonsense reasoning.

\paragraph{Tracking States within Conversations.}
We opt for \textsc{grice} and \textsc{cicero} as the conversation structure they encode, contains frequent references to objects, events or people which change locations or states.
The capability of LMs to keep track of complex states has been a subject of a number of works \citep{prakash2024fine, puerto2024code,Kim2024CodePI}, which have revealed that most models struggle with memorizing, and updating, states of entities mentioned in context.
Within our work, we evaluate conversational elements as well as entity tracking by manually and automatically introducing local alterations to conversational contexts which alter its semantics. In this way, we evaluate the robustness of LMs performance on such tasks, providing stronger guarantees for their practical usage.



\section{Robustness of LM Representations Under Lexically Minimal Alterations}
\label{sec:robustness_formulation}

In our work, we evaluate the effect of minimal alterations to dialogue-based contexts on the precision and stability of LMs latent representations, and in turn on their capability to accurately answer questions.
Here, we outline the key properties of LMs targeted by our alteration-based benchmark.
\paragraph{Problem Definition.} Let $x \in \mathcal{X}$ represent a conversational context consisting of a sequence of utterance pairs $u_i$ between two speakers, $q \in \mathcal{Q}$ be a question posed about this conversation.
We study the capability of the LM as a function $f: \mathcal{X} \times  \mathcal{Q} \to \mathcal{Y}$, where $\mathcal{Y}$ is the space of possible answers. In our work, we focus on yes-no questions, so $\mathcal{Y} = \{\text{yes}, \text{no}\}$.
Let us, without loss of generality, separate $f$ into two components: the encoding of $q, x$ into the latent space $h = e(q,x)$ and answer generation based on the latent representation $y = d(h)$ (decoding).
In our work, we consider how each Transformer layer encodes the conversation, $h_c = e(x) = [h_l]_{\ell=1}^L$. 


Consider an alteration function $\delta: \mathcal{X} \times \mathcal{Q} \to \mathcal{X} \times \mathcal{Q}$, which introduces minimal changes to the original conversation and question.
The altered input is then given as $\hat{x}, \hat{q} = \delta(x, q)$ and its corresponding latent space encoding and answers are $\hat{h} =[\hat{h}_l]_{l=1}^L$ and $\hat{y}$.
We design the alterations to differ minimally in terms of context tokens, while still causing a large impact on conversation semantics. 
\paragraph{Injectivity.} Formally, a function $f$ is injective if it maps distinct inputs to distinct outputs $f(x) \neq f(\hat{x}) \,\, \text{when} \,\, x \neq \hat{x}$. 
Given the high dimensionality of the hidden representations of modern LMs, the encoding $e(.)$ is likely to be injective.
In our work, we investigate a \textit{soft} notion of injectivity through representation similarity of the latent conversation encoding $h_c$. Since the same question can be asked for minimally varying contexts, it \textit{should} encode the full conversation context in $h_c$. Therefore, we expect cases where the encoding fails to capture nuances introduced by alterations to have high \underline{similarity} with the base conversation. We measure this by calculating the direct effect, i.e., the \underline{change} \textit{in confidence} through a causal intervention on the original token stream by replacing all ``aligned'' token representations from the altered token stream. But, the effect of the ``altered tokens'' will be present as we are injecting all of the downstream, previous layer's self-attention effects.
Formally, let $(x + q, y^{\rm gold}),\;(\hat{x}+\hat{q}, \hat{y}^{\rm gold}) \;\in\;\mathcal D$ be an original–altered pair in our dataset. 
Let the confidence corresponding to the \textbf{Original Run (OR)}  and \textbf{Altered Run (AR)} be $P\bigl(\hat{y}^{\text{gold}}\mid x + q\bigr)$ and $P\bigl(\hat{y}^{\text{gold}}\mid \hat{x}+\hat{q}\bigr)$ respectively. 
Let $R^{P}_l = \texttt{Residual}_l(P)$ be the layer~$\ell$ residual stream activations produced by a standard forward pass with $P$ as the input prompt. Following \citet{chattopadhyay2019neuralnetworkattributionscausal}, we now define \textbf{Direct Effect} at layer~$\ell$ as the change in answer probability obtained by patching the altered residuals in the original run: 
\begin{align*}
DE(R^{\hat{x}}_\ell \rightarrow R^{x}_\ell) 
&= P\bigl(\hat{y}^{\text{gold}} \mid x+q;\, \texttt{patch}_\ell(R^{\hat{x}}_\ell)\bigr) \\
&\quad - P\bigl(\hat{y}^{\text{gold}} \mid x+q \bigr)
\end{align*}
where, $\texttt{patch}_\ell()$ is the patching operator that replaces layer~$\ell$ residuals with $R^{\hat{x}}_\ell$, and $R^{\hat{x}}_\ell \rightarrow R^{x}_\ell$ indicates that we patch from the altered run into the original run.
%
%
\paragraph{Local World Model Stability.}
LMs need to keep track of the rich conversational state after each utterance to recover relevant information.
Following work that shows LMs decodably encode world states \citep{toshniwal2022chess,karvonen2024emergent}, we hypothesize the same holds for conversational contexts, where LM layers encode conversation states.
We evaluate the capacity of the model to encode altered inputs (through direct effect analysis), decodability of the information from the world model (through robust accuracy), and estimate the stability of the internal layer-wise representation encoded by the model (direct effect analysis and MLP zero-out ablation). 
Formally, let \( P^{(\mathrm{MLP}_\ell \,\rightarrow\, 0)}(y \mid x+q) \) denote the model’s predicted probability when the MLP submodule at layer \( \ell \) is zeroed out, and \( P(y \mid x+q) \) denote the probability when no such intervention is applied. Then the predicted label for the case of MLP zero-out ablation is given by:
$
\hat{y}
\;=\;
\arg\max_{\text{y}\in\{\texttt{yes},\texttt{no}\}}
\;P^{(\mathrm{MLP}_\ell \,\rightarrow \,0)}\bigl(y \mid x+q\bigr)
$\\
Let $A_\ell$ denotes the accuracy over the entire dataset $\mathcal{D}$ after zeroing out MLP submodule at layer~$\ell$, and $A_0$ be the accuracy when no such intervention is applied. We classify layer \(\ell\) as \textit{useful} if $A_\ell < A_0$, and \textit{harmful} if $A_\ell > A_0$.


\section{Dataset Curation}

We source \datasetname{} from \textsc{grice} \citep{zheng-etal-2021-grice} and \textsc{cicero} \citep{ghosal-etal-2022-cicero}.
These datasets are conversational question answering datasets where the conversation determines a ``local'' context, or world, in the form of an alternating conversation between two people/agents, followed by a probative yes/no question about the state of entities or agents in this world.\footnote{We specifically filter the instances in both of these datasets to only include questions with yes/no answers.} 
To probe if the LM's world models in conversational contexts are malleable, we devise minimal lexical alterations that manipulate entities or perform local semantic changes (e.g., negating actions or statements). 
If the LM has a sufficiently malleable world model, it should accurately answer questions pertaining to both the original and altered instances. 

\paragraph{Selection of Seed Conversational Data.}
\label{sec:seed-selection}
We manually select $44$ representative seed conversations from \textsc{grice} and $33$ from \textsc{cicero}, respectively.
We show an example of seed conversation from both the \textsc{grice} and \textsc{cicero} datasets in the Supplementary material Table~\ref{tab:seedconversations}.

\paragraph{Linguistic Alteration Categories.}
\label{sec:perturb}

We devise the alterations in such a way that they introduce  \textit{minimal} lexical changes which strongly affect the local world model and the semantics of the utterances.
Such alterations will evaluate the capability of LMs to track and update states of entities throughout the conversation.
We manually or semi-automatically apply the following $7$ lexical changes when possible to the seed conversations. 
\\
\textbf{1) Negation.}~~ We negate auxiliary verbs such as ``are'', ``aren't'', ``did'', etc., in a way that changes the truth value of a proposition in the conversational context.
E.g., if the original conversation has a pair of turns: ``A: Did Noah leave oranges in the garage, B: he \underline{didn't}'' $\Rightarrow$ `A: Did Noah leave oranges in the garage, B: he \underline{did}'', making it so that at least some oranges can be found in the garage.\\
\textbf{2) Variable Substitution.}~~ We substitute an entity or object in a conversational context with another one.
E.g. ``A: Did you leave \underline{oranges} in the garage?, B: I left the \underline{watermelons} there.'' $\Rightarrow$ ``A: Did you leave the \underline{oranges} in the garage?, B: I left the \underline{oranges} there.''.\\
\textbf{3) Quantity change.}~~ We change the quantity of countable noun entities.
E.g., ``A: Are some strawberries in the backyard?, B: \underline{Two} are there.'' $\Rightarrow$ ``A: Are some strawberries in the backyard?, B: \underline{Three} are there.''.\\
\textbf{4) Variable Swap.}~~ We swap an entity or object in a conversational context with another one.
E.g. ``A: Did you leave \underline{oranges} in the garage?, B: I left the \underline{watermelons} there.'' $\Rightarrow$ ``A: Did you leave the \underline{watermelons} in the garage?, B: I left the \underline{oranges} there.''\\
\textbf{5) Quantifier change.}~~ We manipulate a quantifier determiner, modifying an entity in a way that alters the implied quantity of the entity.
E.g., ``B: \underline{All} grapefruits are in the playroom.'' $\Rightarrow$ ``B: \underline{Some} grapefruits are in the playroom.''.
Importantly, note that the altered turn is technically implied by the original turn; however, due to scalar implicature \citep{carston1998informativeness} ``some'' is interpreted as ``not all'' assuming that the speakers in the context are following the cooperative principle.\\
\textbf{6) Logical Connective Change.}~~ We manipulate a conjunction combining multiple propositions.
E.g., ``B: Jack put apples \underline{and} oranges in the backyard'' $\Rightarrow$ ``B: Jack put apples \underline{or} oranges in the backyard.''.
This alteration changes the conversational turn from an assertion of two propositions to an uncertain expression of either one of them being \citep[exclusively;][]{exclusive_OR_NL_jennings1994genealogy} true.\\
\textbf{7) Injecting Inconsistent Data.}~~
To ascertain whether models utilize their commonsense knowledge, we inject plausible information that may contradict with common sense.
E.g., ``	B: And a big birthday cake too, with fifty candles..''  $\Rightarrow$ ``And a birthday cake \underline{that changes color every time someone claps}, with fifty candles''.

\paragraph{Manual Dataset Construction.}
\label{sec:manual-data}

We manually apply these linguistic alterations to create $500$ ($300$ from GRICE and $200$ from CICERO) conversations from the $77$ seed conversations. 
We refer to this as the \textit{manual} split of the \textsc{PragWorld} dataset. Each example is annotated by one author and then reviewed by another author to check whether (a) the alterations are lexically minimal and meaningful, (b) ground truth answers remain logically consistent with the altered context and (c) ambiguities and inconsistencies do not create nonsensical inputs. 
We include an example in the final version of our benchmarks only if both authors agree. 

\begin{figure}[t]
    \centering
    \includegraphics[width=\linewidth]{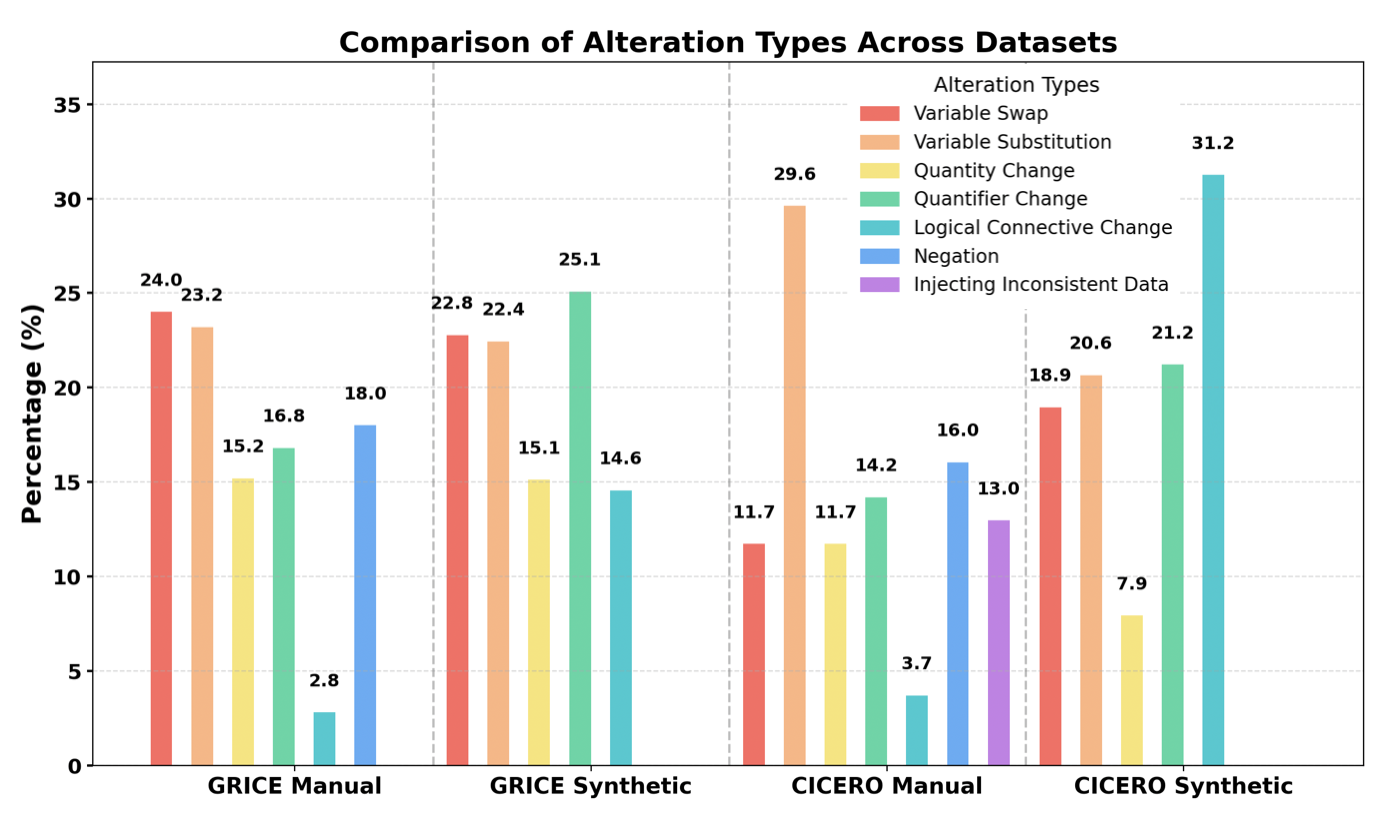}
    \caption{Percentage of conversations with each linguistic alteration category for both splits of \datasetname{}.}
    \label{fig:perturbation_comparison}
\end{figure}

\paragraph{Synthetic Dataset Generation.}
\label{sec:synth-data}

We generate the synthetic dataset split using a semi-automatic pipeline with the following steps. First, we take a set of $104$ initial seed conversation ($49$ from GRICE, and $55$ from CICERO).
Using these, we prompt \texttt{GPT-4-turbo-2024-04-09} \citep{openai2024gpt4technicalreport} to generate novel conversations similar to the seed conversations. The prompts used for the \texttt{GPT-4-turbo-2024-04-09} are shown in Supplementary Section~F.
Secondly, we apply minimal lexical changes to these newly generated conversations using deterministic algorithms (Algorithms 1-5 in Supplementary Section A.1) for each alteration.
In the third step, we create predefined questions automatically using templates (Supplementary Section A.2) for each conversation.
We consider three types of questions: a) \textbf{Quantity Questions}: Ask about a specific quantity of an object. e.g. "Did Lucas place \textbf{six} apples in the kitchen?", b) \textbf{Universal Quantifier questions}: Ask whether a statement applies to all members of a group. e.g. "Did Lucas place \textbf{all} apples in the kitchen?", c) \textbf{Existential Quantifier questions}: Ask whether a statement applies to at least one member of a group. e.g. "Did Lucas place \textbf{some} apples in the kitchen?".
The answer to all the categories of questions can either be \textbf{Yes} or \textbf{No}.
Finally, we manually annotate answers to questions created in the previous step to ensure correctness. 
\label{sec:stats}
\paragraph{Dataset Statistics}
The manual split of the \textsc{PragWorld} consists of a total of $500$ altered conversations, while the synthetic split consists of $2114$ such conversations.
Among the latter, $1074$ instances have the answer ``Yes'' to the question, while $1040$ instances have the answer ``No''. Figure~\ref{fig:perturbation_comparison}, and Supplementary Figure~\ref{fig:length_distribution2} shows the distribution of conversations with respect to the alteration categories and context length, respectively.


\section{Robustness of Internal Representations} 
\label{sec:experiments}

\begin{table*}[!htb]
\resizebox{\textwidth}{!}{%
\begin{tabular}{@{}lcclllllll@{}}
\toprule
 & \multicolumn{1}{l}{\textbf{\#Params}} & \multicolumn{1}{l}{\textbf{Context}} & \textbf{Robust Acc} & \textbf{Yes Acc} & \textbf{No Acc} & \textbf{Original Acc} & \textbf{Altered Acc} & \textbf{Flip Acc} & \textbf{Invariant Acc} \\ \midrule
 \multicolumn{10}{c}{\textbf{\textsc{PragWorld} (manual)} } \\ \midrule
GPT-3.5 & -                                    & 16k     &   42.86 & 52.71 & 93.72 & 71.43 & 70.90 & 71.94 & 69.23             \\
DeepSeek-Inst & 16B (2.4B active) & 128k     & 46.94  & \underline{77.26} & 70.85 & 75.51 & \underline{74.13} & \underline{74.82} & 73.63                 \\\midrule
Phi-3-mini-4k-ins.                   & 3.8B                                 & 4k                                        &    47.96 & 55.60 & 96.86 & 74.49 & 73.88 & \underline{74.82} & 72.53             \\
Phi-3.5-mini-ins.                    & 3.8B (active)                        & 128k                                      &      \underline{48.98} & 66.06 & 86.10 & \underline{78.57} & \underline{74.13} & 74.10 & \underline{74.36}                  \\\midrule

Llama-3.2-1B-Ins.                              & 1B                                   & 128k                                      &   14.29 & 10.83 & 95.96 & 48.98 & 48.76 & 55.4 & 45.05                    \\
Llama-3.2-3B-Ins.                              & 3B                                   & 128k                                      &     20.41 & 16.97 & \textbf{98.65} & 54.08 & 53.23 & 63.31 & 48.35                  \\
Llama-3.1-8B-Ins.                              & 8B                                   & 128k                                      &  \underline{48.98} & 54.87 & 94.62 & 74.49 & 72.14 & \underline{74.82} & 70.33
\\\midrule
Qwen2.5-0.5B-Ins.                         & 500M                                 & 128k                                      &    19.39 & 57.04 & 54.71 & 48.98 & 57.71 & 69.78 & 50.92                    \\
Qwen2.5-1.5B-Ins. & 1.5B         & 128k              &                           22.45 & \textbf{93.50} & 28.25 & 65.31 & 64.18 & 56.12 & 68.50             \\
Qwen2.5-7B-Ins.                           & 7B                                   & 128k                                      &   37.76 & 47.65 & 95.96 & 68.37 & 69.40 & 72.66 & 66.67               \\  \bottomrule
 \multicolumn{10}{c}{\textbf{\textsc{PragWorld} (synthetic)} } \\ \midrule
GPT-3.5 & -                                    & 16k     &   67.21 & 80.63 & 91.44 & \textbf{87.04} & \textbf{83.35} & \textbf{82.85} & \textbf{87.38}               \\
DeepSeek-Inst & 16B (2.4B active) & 128k     & 60.93  & \underline{93.95} & 64.13 & 80.97 & 76.68 & 76.25 & 81.00                   \\\midrule
Phi-3-mini-4k-ins.                   & 3.8B                                 & 4k                                        &    \textbf{64.78} & 79.14 & 87.02 & \underline{84.62} & \underline{80.35} & \underline{80.15} & \underline{84.11}                    \\
Phi-3.5-mini-ins.                    & 3.8B (active)                        & 128k                                      &      \underline{63.97} & 90.69 & 70.38 & 83.60 & 77.70 & 78.10 & 81.31                  \\\midrule

Llama-3.2-1B-Ins.                              & 1B                                   & 128k                                      &   47.77 & 22.16 & \underline{95.10} & 58.91 & 56.25 & 57.12 & 56.23                    \\
Llama-3.2-3B-Ins.                              & 3B                                   & 128k                                      &     47.98 & 24.39 & \textbf{96.63} & 61.13 & 57.99 & 58.58 & 59.03                      \\
Llama-3.1-8B-Ins.                              & 8B                                   & 128k                                      &  60.93 & 63.69 & 94.90 & 80.77 & 76.44 & 76.45 & 79.75                  \\\midrule
Qwen2.5-0.5B-Ins.                         & 500M                                 & 128k                                      &    58.50 & 70.30 & 80.96 & 75.91 & 73.44 & 72.89 & 76.64                    \\
Qwen2.5-1.5B-Ins. & 1.5B         & 128k              &                           55.87 & \textbf{96.28} & 45.67 & 74.09 & 68.69 & 68.93 & 72.27             \\
Qwen2.5-7B-Ins.                           & 7B                                   & 128k                                      &   60.73 & 70.39 & 90.96 & 82.19 & 77.88 & 78.30 & 80.22               \\  \bottomrule
\end{tabular}%
}
\caption{Model performance on \datasetname{}. Deepseek-Inst: Deepseek-coder-V2-Lite-Instruct, GPT-3.5: GPT-3.5-Turbo (Dec 2023). Best and second best performances are represented by \textit{bold}, and \textit{underline} respectively.}
\label{tab:globalresultstable}
\end{table*}

In this section, we report performance of various open- and closed-source LMs on both splits of \datasetname{} and perform various analyses showcasing which issues present in LMs our benchmark targets.

\paragraph{Baselines and Performance Metrics.}
We evaluate both closed-source models, such as \texttt{GPT-3.5-Turbo}, and open-source models, including \texttt{Deepseek-coder-instruct} (16B), the \texttt{Phi-3} series, the \texttt{Llama-3} series, and the \texttt{Qwen2.5} series. We provide details of the models we use in Supplementary Table \ref{tab:baseline_models} and the prompts used in the Supplementary Table \ref{tab:inference_prompts}.
Throughout our experiments, we report a number of individual and aggregate metrics.
\textbf{Robust accuracy} deems a model response to be correct if and only if it correctly answers both the original conversation and all of its altered instances.
We additionally report accuracy across different alteration subsets: \textbf{Flip Accuracy} (accuracy on alterations designed to change the answer), \textbf{Invariant Accuracy} (accuracy on alterations designed to maintain the answer), \textbf{Original Accuracy} (accuracy on unaltered conversations), and \textbf{Altered Accuracy} (accuracy on altered conversations).
We also report individual accuracies for each gold label -- "Yes" (\textbf{Yes Accuracy}) or "No" (\textbf{No Accuracy}).

\subsection{Performance on \datasetname}
\label{sec:exp:perf}

\begin{table*}[!htb]
\resizebox{\textwidth}{!}{%
\begin{tabular}{@{}lcllllllll@{}}
\toprule
 & \multicolumn{1}{l}{\textbf{\#Params}} & \multicolumn{1}{l}{\textbf{Context Size}} & \textbf{Robust Acc} & \textbf{Yes Acc} & \textbf{No Acc} & \textbf{Original Acc} & \textbf{Altered Acc} & \textbf{Flip Acc} & \textbf{Invariant Acc} \\ \midrule
 \multicolumn{10}{c}{\textbf{\textsc{PragWorld} (manual)} } \\ \midrule
 Phi-3-mini-4k-ins.                              & 3.8B                                   & 4k                                      &  50.00\textcolor{gray}{(+2.04)}	& 66.43	& \textbf{91.03}	& 78.57	&77.11 	& 75.54	&  \underline{77.29}                \\ 
Phi-3.5-mini-ins.                    & 3.8B (active)                        & 128k                                      &      52.04 \textcolor{gray}{(+3.06)}                       &  70.76              &    86.55             &   \underline{81.63}                  &    76.87                    & 76.98              &     76.56                  \\ \midrule 
Llama-3.2-1B-Ins.                              & 1B                                   & 128k                                      &  32.65 \textcolor{gray}{(+18.36)} 	&	59.21 &	78.92 &	67.35 &68.16	& 68.35&	68.50                   \\
Llama-3.2-3B-Ins.                              & 3B                                   & 128k                                      &  48.98\textcolor{gray}{(+28.57)} 	&	63.9 &	\underline{87.44} &	77.55 &73.63	& \underline{79.14}&	70.70                  \\
Llama-3.1-8B-Ins.                              & 8B                                   & 128k                                      &  \textbf{59.18}\textcolor{gray}{(+10.2)} 	& \textbf{79.06}	& 84.3	& 	\textbf{87.76}& \textbf{79.85}	& 76.98	&   \textbf{81.68}               \\ \midrule
Qwen2.5-0.5B-Ins. & 500M        & 128k              &     22.45\textcolor{gray}{(+3.06)}                                                 &    44.77            &   73.09                   &  60.20                     &  56.72                &  66.91    &    52.01           \\
Qwen2.5-1.5B-Ins. & 1.5B         & 128k              &     47.96\textcolor{gray}{(+25.51)}                                                 &    \underline{71.12}             &   76.23                   &  74.49                      &  73.13                &  77.70     &    71.06           \\
Qwen2.5-7B-Ins.                           & 7B                                   & 128k                                      &    \underline{55.10}\textcolor{gray}{(+17.34)}                         &  70.04              &     \textbf{91.03}           &  80.61                    &  \underline{79.10}                     &  \textbf{82.01}                &   76.92                    \\ \bottomrule
\end{tabular}%
}
\caption{Performance of best performing LMs after fine-tuning on the synthetic split. The relative improvement in robust accuracy over the non-finetuned counterparts given in \textcolor{gray}{\textbf{gray}}.}
\label{tab:finetunedtable}
\end{table*}

We report performance of analyzed models on the manual and synthetic splits of \datasetname{} in \Cref{tab:globalresultstable}. 
Overall, the \texttt{Phi} series models are the best performing ones according to robust accuracy, surpassing even larger models like \texttt{GPT-3.5}.
Most models report a large gap between \textbf{Yes} and \textbf{No} accuracy on both splits, indicating their preference towards one of the answer options.
This disparity is especially pronounced in smaller models like \texttt{Llama-3.2-1B-Ins.},
\texttt{Llama-3.2-3B-Ins.}, 
and \texttt{Qwen-2.5-1.5B-Ins.}
\texttt{GPT-3.5}
and \texttt{DeepSeek-Inst.}
also exhibit this bias, more frequently answering \textbf{No}.

\paragraph{Fine-tuning on Synthetic Data.} To improve robust accuracy and reduce the performance discrepancies between Yes and No accuracy, we fine-tune the models on the synthetic data split and evaluate them on the manual split. 
We report performance of fine-tuned models on the manual split of \textsc{PragWorld} in \Cref{tab:finetunedtable}.
All tuned models show improvements in Robust Accuracy, with notable gains for the \texttt{Llama-3} and \texttt{Qwen-2.5} series.
Smaller models like \texttt{Llama-3.2-1B-Ins.} and \texttt{Qwen-2.5-1.5B-Ins.} also show large relative improvements in Robust Accuracy, indicating that fine-tuning on synthetic data is particularly beneficial for smaller models. 
These results suggest that the synthetic split encodes valuable training signal that helps models better handle input alterations.
Despite these improvements, the persistent gap between Yes and No Accuracy in some models suggests that fine-tuning alone does not mitigate inherent biases in LMs.

\begin{table}[!htpb]
\resizebox{0.46\textwidth}{!}{%
\begin{tabular}{@{}l|ccc|ccc}
\toprule
\multicolumn{7}{c}{\textbf{\textsc{PragWorld} (manual)}} \\ 
\midrule
 & \multicolumn{3}{c|}{\textbf{Base Model}} & \multicolumn{3}{c}{\textbf{Fine-tuned Model}} \\ 
 & \textbf{Yes Acc.} & \textbf{No Acc.} & \textbf{Total Acc.} & \textbf{Yes Acc.} & \textbf{No Acc.} & \textbf{Total Acc.} \\ 
\midrule
GPT-3.5 
& 79.12 & 92.85 & \textbf{85.98} & - & - & - \\  
Deepseek-inst 
& \underline{81.86} & 79.12 & 80.49 & - & - & - \\  
Phi-3-mini-4k-ins. 
& 55.49 & \underline{96.70} & 76.09 & 71.42 & \textbf{95.05} & \underline{83.24} \\  
Phi-3.5-mini-ins.  
& 78.57 & 84.61 & 81.59 & 70.87& \underline{94.50} & 82.69 \\  
\midrule
Llama-3.2-1B-Ins. 
& 6.04 & \textbf{98.35} & 52.19 & 53.29 & 90.65 & 71.97 \\  
Llama-3.2-3B-Ins. 
& 39.01 & 93.40 & 66.20 & 70.87 & 90.65 & 80.76 \\  
Llama-3.1-8B-Ins. 
& \underline{81.86} & 88.46 & \underline{85.16} & \textbf{84.61} & 91.20 & \textbf{87.91} \\  
\midrule
Qwen2.5-0.5B-Ins. 
& \textbf{100.00} & 3.29 & 51.64 & 56.04 & 80.21 & 68.13 \\  
Qwen2.5-1.5B-Ins. 
& 80.21 & 65.38 & 72.80 & \underline{74.17} & 85.71 & 79.94 \\  
Qwen2.5-7B-Ins. 
& 63.73 & 93.95 & 78.84 & 73.07 & 92.85 & 82.96 \\  

\bottomrule
\end{tabular}%
}
\caption{Results of the probing experiment regarding the capability of LMs to memorize information from conversations. Generally, models are able to recall most of the information presented in conversational context. We find that this capability also improves with scale.}
\label{tab:entitytrackingtable}
\end{table}


\subsection{Benchmarking Entity Tracking Ability of LMs}
\label{sec:exp:entitytracking}
To evaluate whether LMs are able to accurately keep track of and update the local world state throughout the conversation, we conduct the following experiment.
We segment the original conversations (from the manual split of \textsc{PragWorld}) after some number of utterances, generating progressively longer alterations or parts of the original conversation. We select $15$ original conversations and partition them into three or four groups based on the dataset they were seeded from.
For each of these alterations, we create an equal proportion of Yes/No questions regarding entities and people mentioned in the conversation.
We then probe the models with these questions in order to determine whether they accurately memorize information from the conversational context. 
The final dataset for the entity tracking experiment contains $364$ instances with a balanced proportion of "Yes" and "No" answers. 

We report model performance on the entity tracking experiment in \Cref{tab:entitytrackingtable}. We see that fine-tuning consistently improves entity tracking performance across all models, especially models of smaller sizes like \texttt{Llama-3.2-1B-Ins.}, and \texttt{Qwen2.5-0.5B-Ins.}
Larger base models like \texttt{Llama-3.1-8B-Ins.}, and \texttt{Qwen2.5-7B-Ins.} initially perform well, indicating that the capacity for entity tracking is present within the models.
Overall, our findings suggest that fine-tuning and model scale play a crucial role in enhancing entity tracking. 

\subsection{Effect of Conversation Length and Answer Type}
\label{sec:exp:indepth}

We observe that the performance gap between Yes and No accuracy is much more pronounced in the manual split compared to the synthetic one. 
Supplementary Table \ref{tab:accuracy_distribution} shows the distribution of Yes and No accuracy of \texttt{GPT-3.5-Turbo} w.r.t alteration categories for both splits of \datasetname{}.
One possible explanation is that the manual split contains a greater variety of questions compared to the synthetic one.
We find that \texttt{GPT-3.5-Turbo} particularly struggles with two specific question varieties which are not present in the synthetic split due to the difficulty of automatically generating such variations, as seen in Supplementary Table \ref{tab:question_distribution-fractional-fromat}.

We also analyze the effect of conversation length, in the number of utterances, on accuracy.
We bucket conversations into three categories: \textsc{Short}, with less than 11 utterances, \textsc{Medium}, between 11 to 15 utterances, and \textsc{Long}, with conversations longer than 15 utterances. Supplementary Table \ref{tab:accuracy_distribution_across_length} shows that accuracy of \texttt{GPT-3.5} increases from \textsc{Short} to \textsc{Medium} conversations, and then decreases for \textsc{Long} conversations, except for ''No'' accuracy in Synthetic split. 
It shows that accuracy is affected by many factors such as length, Type of Questions, Model type, and type of perturbation. 
We investigate the confounding effects of pragmatic phenomena in Supplementary Section J.

\section{Insights from a Dual-Perspective Mechanistic Interpretability framework }
To analyze internal representations, we perform \textit{causal interventions} at the residual stream, and MLP sublayers, resulting in two techniques: 1) Direct Effect Patching, 2) MLP Zero-Out ablation (introduced before). We follow \citet{joshi-etal-2025-towards} and \citet{geva2023dissectingrecallfactualassociations} respectively, but explain the differences in Supplementary section G.

\begin{figure}[!ht]
    \centering
    \begin{subfigure}[t]{0.5\textwidth}
        \centering
    \includegraphics[width=0.9\linewidth]{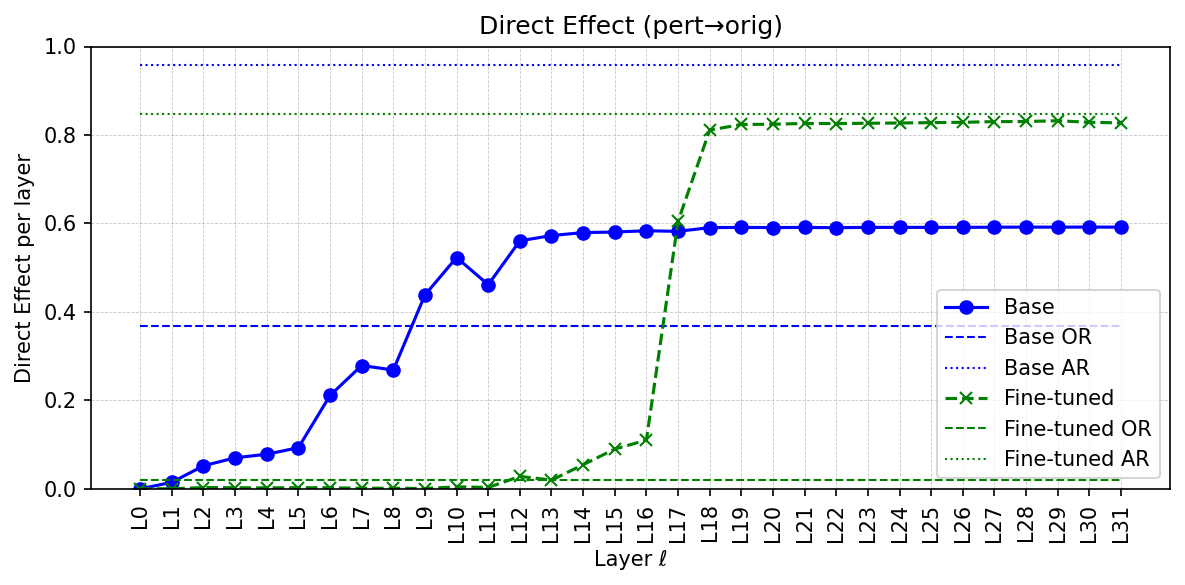}  \caption{}
    \label{fig:DE_Phi_correct_raw_ftuned}
    \end{subfigure}
    \begin{subfigure}[t]{0.5\textwidth}
        \centering
    \includegraphics[width=0.9\linewidth]{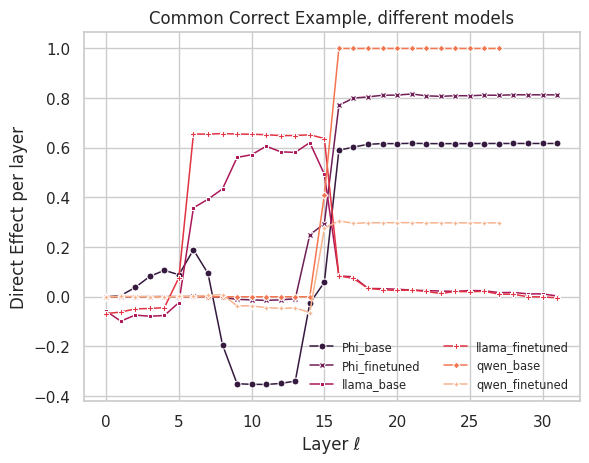}  \caption{}
    \label{fig:DE_common_examples}
    \end{subfigure}
    \caption{(a) Comparison of the Direct Effect Probing the base and fine-tuned versions of the \texttt{Phi-3.5-mini-instruct} for an example where both the base and fine-tuned versions are correct on both the original and altered conversation. 
    (b) DE patching for three models on a common example, where the base and finetuned models are correct on original and altered instances.}
    \label{fig:DE_combined}
\end{figure}

\subsection{Results of Direct Effect Patching}
We now study the base and fine-tuned variants of the \texttt{Phi-3.5-mini-instruct} model.
We first choose an original example and a corresponding altered variant, where both the base and finetuned models predicts the label(s) \textit{correctly}. Fig.~\ref{fig:DE_Phi_correct_raw_ftuned} shows that for the chosen \textit{correct} example, fine-tuned model confidence for the correct altered label increases sharply. 
The fine-tuned model is robust to patching up to layer 16, but its confidence changes dramatically as we move from layer 16 to 17. 
In addition, the fine-tuned version shows a low \textbf{OR}, and high \textbf{AR} confidence as compared to the base version. 
We observe a similar effect for a sample where the base model is incorrect, while the fine-tuned version is correct on both the original and altered conversation, shown in Figure \ref{fig:DE_Phi_incorrect_raw_ftuned} of Supplementary. 
We expand our analysis to also include \texttt{Llama-3.1-8B-Ins}, and \texttt{Qwen2.5-7B-Ins} and compare direct effect patching for the same instance where all models are correct in \Cref{fig:DE_common_examples}.
The base and fine-tuned version of Phi and Qwen show an increase in  the correct altered label confidence. Their fine-tuned versions also show robustness towards alterations in early layers, and then a sudden increase in confidence between layers 14 and 15. For both Llama versions, we observe an opposite trend where the initial layers show a sudden increase in confidence (at layer 5), which decreases drastically moving from layer 15 to 16. 

\subsection{Results of MLP zero-out Ablation}
\begin{figure}[!htb]
    \centering
\includegraphics[width=1\linewidth]{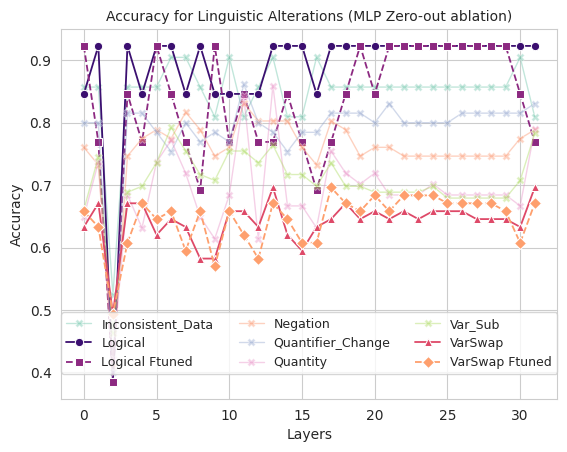}  
    \caption{Effect of MLP zero-out ablation on the accuracies (dataset-wide) for different alteration types of base \texttt{Phi-3.5-mini-ins}. For \textit{Logical Connective}, and \textit{Variable Swap}, we also report results for the fine-tuned model. Other alteration types are transparent for emphasis. 
    }
    \label{fig:MLP_ablation_Phi_raw_finetuend_combined_pcategories}
\end{figure}

\begin{figure}[t]
    \centering
\includegraphics[width=1.0\linewidth]{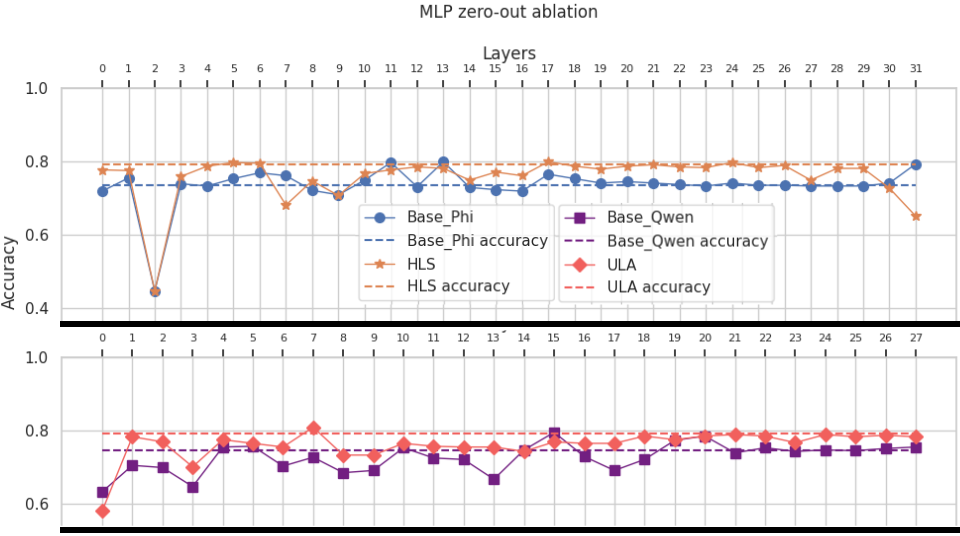}  
    \caption{Results of MLP zero-out ablation on \texttt{Phi-3.5-mini-ins}, and \texttt{Qwen2.5-7B-Ins} fine-tuned using HLS and ULA. Base accuracy denotes the base model accuracy on the entire dataset, and HLS/ULA accuracy is the accuracy after finetuning using HLS/ULA.}
    \label{fig:MLP_ablation_Phi_raw_harmfulsuppresssion}
\end{figure}

We report the results of \textit{MLP zero-out ablation} for Phi-3.5 (base) in \Cref{fig:MLP_ablation_Phi_raw_harmfulsuppresssion}. 
We find (\textit{blue line}) that \textit{MLP zero-out} at layers $2, 9,$ and $16$ leads to a decrease in accuracy indicating that these layers are \textit{useful}.
Ablating layers $5, 6, 7, 11, 13, 17,$ and $31$ results in improved accuracy, suggesting that these are \textit{harmful} containing spurious signals or shortcut patterns. 
We investigate the effect of MLP zero-out ablation on different linguistic alterations. As shown in \Cref{fig:MLP_ablation_Phi_raw_finetuend_combined_pcategories}, \textit{Logical Connective} is most impacted by harmful layers, while \textit{Variable Swap} is the least affected. We also find that these two cases are less influenced by harmful layers in the fine-tuned version of the same model in the same figure. 
\textit{Logical Connective} is less affected in the early layers of the fine-tuned version compared to the base version, while the accuracy decreases in the final layers after ablating. The latter also holds for \textit{Variable Swap}. 
Hence, we find fine-tuning helps suppress the effect of the harmful layers of the Phi model.

\subsection{Effect of additional Layer Regularization}
We design two regularization strategies based on previous observations: a) \textit{Useful Layer Amplification}, and b) \textit{Harmful Layer Suppression}. In \textit{Useful Layer Amplification} (ULA), we choose a set of useful layers and attach a small two‐layer classification head on top of every layer’s MLP output. 
The ULA loss is the mean of classification losses in all useful layers, which we add to the next-token prediction loss, weighted by a \textit{useful weight} $\alpha \in [0,1]$.
In \textit{Harmful Layer Suppression} (HLS), we choose a set of harmful layers, and add an $\mathrm{L}_2$ penalty on MLP’s output (the residual output after the feed‐forward submodule) of each such layer. The final loss is the mean squared norm of those activations, which is then added to the model's next-token prediction loss, weighted by a \textit{harmful weight} $\beta \in [0, 1]$
\Cref{fig:MLP_ablation_Phi_raw_harmfulsuppresssion} reports the effect of HLS (\textit{orange line}) (for harmful layers $7, 11, 13, 17,$  and $31$ visible in Base\_Phi with \textit{harmful weight} = $1.0e-3$) on \texttt{Phi-3.5-mini-instruct}. 
The same figure reports the harmful and useful layers for the base version of the model \texttt{Qwen2.5-7B-Instruct}, and the effect of ULA strategy (\textit{red line}) (useful layers $0$, $3$, and $13$ visible in Base\_Qwen with \textit{useful weight} = 1.0e-3), where accuracy decreases for both useful layers and harmful layers. All layers show accuracy lower than \textit{ULA accuracy}.

\section{Conclusion}

We introduced seven types of linguistic alterations to two dyadic conversation datasets (\textsc{grice} and \textsc{cicero}), creating the \textsc{PragWorld} benchmark with the goal to assess the malleability of LLMs' \textit{implicit} world model on these minimal alterations through the lens of robust accuracy. We benchmark open and closed source models, and observe that they are not robust to linguistic alterations, exhibiting substantial memorization errors. Using mechanistic interpretability techniques, we pinpoint \textit{useful} and \textit{harmful} layers, and highlight linguistic alterations most affected by the harmful layers. Inspired by these observations, we propose a two regularization techniques that help suppress the effect of harmful layers. 

\appendix
\section{Semi-Automatic Data Generation Pipeline}
\subsection{Alteration Algorithms}
\label{app:semi-automatic-perturbation-algo}
As described in paragraph \textbf{Synthetic Dataset Generation.} of the section \textbf{Dataset Curation}, the synthetic data is generated by providing GPT-4 with seed conversations from the GRICE dataset and a prompt instructing the model to create new synthetic conversations. The prompt contains three tasks for GPT-4: (1) Identify all agents, objects, and locations in the input seed conversation. (2) Choose two tuples of agents, objects, and locations, i.e. (A1, O1, L1) and (A2, O2, L2)., and (3) Create a conversation using the agents, objects, and locations identified in step 1 with a special focus/ emphasis on (A1, O1, L1), and (A2, O2, L2). Let us assume that for a given conversation, \texttt{AllAgent}, \texttt{AllObject}, and \texttt{AllLocation} represent the lists that contain all the agents, objects, and locations, respectively, extracted from the given conversation. Similarly, \texttt{FocusAgent}, \texttt{FocusObject}, and \texttt{FocusLocation} represent the lists that contain Agents (A1, A2), Objects (O1, O2), and Location (L1, L2), respectively, on which GPT-4 has to pay special focus while generating the given conversation. Let \texttt{RemainingAgent} be \texttt{AllAgent} - \texttt{FocusAgent}, 
\texttt{RemainingObject} be \texttt{AllObject} - \texttt{FocusObject}, and
\texttt{RemainingLocation} be \texttt{AllLocation} - \texttt{FocusLocation}. This section presents the algorithms for the following alteration categories: Variable Swap (\Cref{algo:variable_swap}), Variable Substitution (\Cref{algo:variable_substitution}), Quantity change (\Cref{algo:quantity_change}), Quantifier change (\Cref{algo:quantifier_change}), and Logical Connective change (\Cref{algo:logical_connective}).

\begin{algorithm}[t]
\SetAlgoLined
\textbf{Example:}
\[
\begin{array}{c}
\text{Are all the pumpkins in the \textbf{Kitchen}?} \\
\text{He was arranging the \textbf{living room}.} \\
\Downarrow \\
\text{Are all the pumpkins in the \textbf{living room}?} \\
\text{He was arranging the \textbf{Kitchen}.}
\end{array}
\]

\rule{\linewidth}{0.4pt}
\textbf{Input:} $\textbf{S} \gets$ Seed Conversation \\
\textbf{Initialize}: Initialize the lists \texttt{FocusAgent}, \texttt{FocusObject}, \texttt{FocusLocation}, \texttt{RemainingAgent}, \texttt{RemainingObject}, and \texttt{RemainingLocation}   \\
\textbf{Output} $\textbf{N} \gets$ New synthetic conversation.
\begin{enumerate}
    \item Randomly select either to alter Agent, Object, or Location. \textit{(Let's say we select Agent.)}
    \item If \texttt{len(RemainingAgent) $\geq$ 2}: Randomly select two elements from the list \texttt{RemainingAgent} and \textbf{Swap} them. Here, the answers to the questions cannot change.
    \item If \texttt{len(RemainingAgent) == 1}: \textbf{Swap} the element present in \texttt{RemainingAgent} with the randomly selected element from the list \texttt{FocusAgent}. Here, the answers to the questions can change.
    \item If \texttt{len(RemainingAgent) == 0} and \texttt{len(FocusAgent) >= 1}: Randomly select two elements from the list \texttt{FocusAgent} and \textbf{Swap} them. Here, the answers to the questions can change.

\end{enumerate}
\caption{Variable Swap}
\label{algo:variable_swap}
\end{algorithm}

\begin{algorithm}[t]
\SetAlgoLined
\textbf{Example:} Are all the pumpkins in the \textbf{playroom}? $\Rightarrow$ Are all the pumpkins in the \textbf{pantry}? 
\rule{\linewidth}{0.4pt}
\textbf{Input:} $\textbf{S} \gets$ Seed Conversation \\
\textbf{Initialize}: Initialize the lists \texttt{FocusAgent}, \texttt{FocusObject}, \texttt{FocusLocation}, \texttt{RemainingAgent}, \texttt{RemainingObject}, and \texttt{RemainingLocation}   \\
\textbf{Output} $\textbf{N} \gets$ New synthetic conversation.
\begin{enumerate}
    \item Randomly select either to alter Agent, Object, or Location. \textit{(Let's say we select Agent.)}
    \item If \texttt{len(RemainingAgent) $\geq$ 2}: Randomly select two elements from the list \texttt{RemainingAgent} and \textbf{Replace} them. Here, the answers to the questions cannot change.
    \item If \texttt{len(RemainingAgent) == 1}: Randomly select an element from the list \texttt{FocusAgent} and \textbf{Replace} it with the element present in \texttt{RemainingAgent}. Here, the answers to the questions can change.
    \item If \texttt{len(RemainingAgent) == 0} and \texttt{len(FocusAgent) >= 1}: Randomly select two elements from the list \texttt{FocusAgent} and \textbf{Replace} them. Here, the answers to the questions can change.

\end{enumerate}
\caption{Variable Substitution}
\label{algo:variable_substitution}
\end{algorithm}

\begin{algorithm}[t]
\SetAlgoLined
\textbf{Example:} There are \textbf{four} apples in the kitchen. $\Rightarrow$ There are \textbf{six} apples in the kitchen. 
\rule{\linewidth}{0.4pt}

\textbf{Input:} $\textbf{S} \gets$ Seed Conversation \\

\textbf{Initialize}: Initialize the lists \texttt{FocusAgent}, \texttt{FocusObject}, \texttt{FocusLocation}, \texttt{RemainingAgent}, \texttt{RemainingObject}, and \texttt{RemainingLocation}.   \\

\textbf{Output}: $\textbf{N} \gets$ New synthetic conversation.

\begin{enumerate}
    \item For every sentence/utterance in the conversation:
    \begin{enumerate}
        \item For every word $W$ in the sentence/utterance:
        \begin{enumerate}
            \item If $W$ represents a quantity:
            \begin{enumerate}
                \item Let \texttt{NumForm} be the numerical form of $W$.
                \item Let \texttt{Index} be the utterance number.
            \end{enumerate}
        \end{enumerate}
        \item Collect all the tuples $(\texttt{Index}, W, \texttt{NumForm})$ in the list \texttt{quantity\_detect}.
    \end{enumerate}
    
    \item \textbf{If} \texttt{len(quantity\_detect) > 0}:
    \begin{enumerate}
        \item Choose a random tuple from \texttt{quantity\_detect}.
        \item \textbf{If} \texttt{NumForm == 1}:
        \begin{enumerate}
            \item \texttt{Operator} $\gets$ “addition”
        \end{enumerate}
        \item \textbf{Else}:
        \begin{enumerate}
            \item \texttt{Operator} is randomly chosen from {“addition”, “subtraction”}.
        \end{enumerate}
    \end{enumerate}
    
    \item \textbf{If} \texttt{Operator == addition}:
    \begin{enumerate}
        \item \texttt{tempNo} $\gets$ Choose a random number from the range $[1, \texttt{NumForm}]$.
        \item \texttt{New\_number} $\gets$ \texttt{NumForm} + \texttt{tempNo}.
    \end{enumerate}
    
    \item \textbf{Else If} \texttt{Operator == subtraction}:
    \begin{enumerate}
        \item \texttt{tempNo} $\gets$ Choose a random number from the range $[1, \texttt{NumForm} - 1]$.
        \item \texttt{New\_number} $\gets$ \texttt{NumForm} - \texttt{tempNo}.
    \end{enumerate}

    \item Replace the word $W$ present in the sentence at \texttt{Index} with \texttt{New\_number}.
\end{enumerate}

\caption{Quantity Change}
\label{algo:quantity_change}
\end{algorithm}

\begin{algorithm}[t]
\SetAlgoLined
\textbf{Example:} Jayden placed \textbf{some} apples in kitchen. $\Rightarrow$ Jayden placed \textbf{all} apples in kitchen.
\rule{\linewidth}{0.4pt}
\textbf{Input:} $\textbf{S} \gets$ Seed Conversation \\
\textbf{Initialize}: Initialize the lists \texttt{FocusAgent}, \texttt{FocusObject}, \texttt{FocusLocation}, \texttt{RemainingAgent}, \texttt{RemainingObject}, and \texttt{RemainingLocation}   \\
\textbf{Output} $\textbf{N} \gets$ New synthetic conversation.
\begin{enumerate}
    \item Maintain a mapping dictionary \texttt{Quant\_Map = \{all: some, All: Some, some: all, Some: All\}}
    \item \textbf{Find} all the sentences in which existential quantifier \texttt{some} and universal quantifier \texttt{all} are present. Create a list \texttt{Quantifier\_Detect} that contains all these sentences. 
    \item If \texttt{len(Quantifier\_Detect) > 0}: Choose a random element/ sentence from \texttt{Quantifier\_Detect} in which 
quantifier \texttt{Q} (say) is present.\\
\texttt{New\_quantifier = Quant\_Map[Q]}\\
\textbf{Replace} \texttt{Q} in the given sentence with \texttt{New\_quantifier}.

\end{enumerate}
\caption{Quantifier Change}
\label{algo:quantifier_change}
\end{algorithm}

\begin{algorithm}[t]
\SetAlgoLined
\textbf{Example:} Apples are in the kitchen \textbf{and} bedroom. $\Rightarrow$ Apples are in the kitchen \textbf{or} bedroom.
\rule{\linewidth}{0.4pt}
\textbf{Input:} $\textbf{S} \gets$ Seed Conversation \\
\textbf{Initialize}: Initialize the lists \texttt{FocusAgent}, \texttt{FocusObject}, \texttt{FocusLocation}, \texttt{RemainingAgent}, \texttt{RemainingObject}, and \texttt{RemainingLocation}   \\
\textbf{Output} $\textbf{N} \gets$ New synthetic conversation.
\begin{enumerate}
    \item Maintain a mapping dictionary \texttt{Log\_Map = \{and: or, And: Or, or: and, Or: And\}}
    \item \textbf{Find} all the sentences in which logical connectives like \texttt{or}, and \texttt{and} are present. Create a list \texttt{Logconn\_Detect} that contains all these sentences. 
    \item If \texttt{len(Logconn\_Detect) > 0}: Choose a random element/ sentence from \texttt{Logconn\_Detect} in which 
Logical connective \texttt{L} (say) is present.\\
\texttt{New\_Logconn = Log\_Map[L]}\\
\textbf{Replace} \texttt{L} in the given sentence with \texttt{New\_Logconn}.

\end{enumerate}
\caption{Logical Connective Change}
\label{algo:logical_connective}
\end{algorithm}

\subsection{Synthetic Questions}
\label{app:semi-automatic-synthetic-questions}
The synthetic questions were created automatically for all the instances belonging to the GRICE version in the synthetic variant of the \textsc{PragWorld} dataset. The questions were created using the following templates: 1) \textbf{Quantity Questions}: Did \texttt{<agent>} place \texttt{<quantity>} \texttt{<object>} in the \texttt{<location>}? Eg. Did \textbf{Lucas} place \textbf{six} \textbf{apples} in the \textbf{kitchen}?, 2) \textbf{Universal Quantifier Questions}: Did \texttt{<agent>} place \texttt{<all quantifier>} \texttt{<object>} in the \texttt{<location>}? Eg. Did \textbf{Lucas} place \textbf{all} \textbf{apples} in the \textbf{kitchen}?, and 3) \textbf{Existential Quantifier Questions}: Did \texttt{<agent>} place \texttt{<some quantifier>} \texttt{<object>} in the \texttt{<location>}? Eg. Did \textbf{Lucas} place \textbf{some} \textbf{apples} in the \textbf{kitchen}?\\
The questions for the manual dataset consist of four more varieties apart from the aforementioned synthetic questions as shown in the Table \ref{tab:question_distribution-fractional-fromat}. These four varieties of questions are:  1) \textbf{Only Entity Ques}: It contains only entity, not any quantity, and quantifier related to that particular entity. Eg. Did Lucas place \textbf{apples} in the kitchen?, 2) \textbf{Sure Ques}: It contains the word “sure”. Eg. is Lucas \textbf{sure} about where the apples are?, 3) \textbf{Know Ques}: It contains the word “know”. Eg. does Lucas \textbf{know} who put the apples in the kitchen?, and 4) \textbf{Other Ques}: It contains those questions that doesn't fall in any of the aforementioned categories. Eg. Is the speaker interested about the mentioned sales job?.

\section{Dataset Statistics and Samples}
\label{app:dataset-samples}
In this section, we provide qualitative and quantitative examples of the introduced dataset. \Cref{tab:examplespragworld} shows data samples sourced from both sub-datasets, while \Cref{app:sub-data-dist} provides statistics of the dataset.

\begin{figure}[htp]
    \centering
    \includegraphics[width=0.8\linewidth]{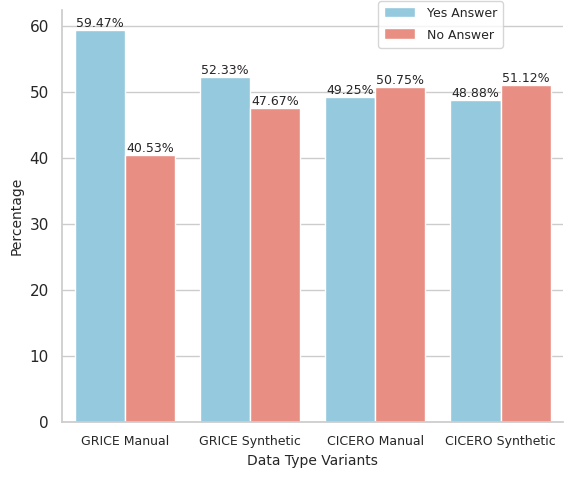}
    \caption{Label distribution: Figure shows the percentage of conversations having questions with gold answer as "Yes" and "No" in both the manual and synthetic variants of the \textsc{PragWorld} dataset.}
    \label{fig:yes_no_acc_finegrained_distribution}
\end{figure}

\begin{figure}[htp]
    \centering
    \includegraphics[width=0.8\linewidth]{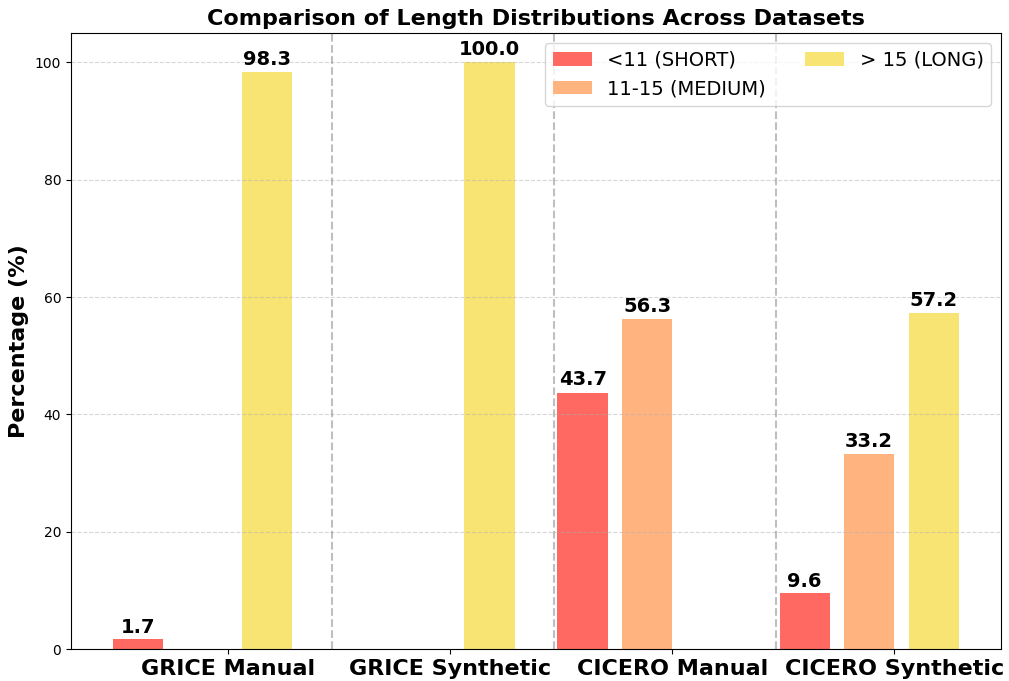}
    \caption{Figure shows the length distribution for both data splits. Here, we consider as short  conversations of length $< 11$, medium corresponds to lengths in the range $[11, 15]$, and long covers conversations of length $> 15$.}
    \label{fig:length_distribution2}
\end{figure}

\begin{table*}[t]
\scriptsize
\resizebox{\textwidth}{!}{%
\begin{tabular}{@{}lll@{}}
\toprule
\multicolumn{1}{c}{\textbf{Conversation}} & \multicolumn{1}{c}{\textbf{Question}} & \textbf{Source} \\
\midrule

\begin{tabular}[t]{@{}l@{}}Alice: are some of the strawberries in the bedroom\\ Bob: three are there\\ Alice: were you there\\ Bob: I was not there\\ Alice: then where\\ Bob: I journeyed to the office and the front\_yard\\ Alice: \colorbox{yellow}{was Jack there}\\ Bob: \colorbox{yellow}{he journeyed to the bedroom}\\ Alice: did he leave the plums in the bedroom\\ Bob: he left them in the bedroom or the front\_yard\\ Alice: \colorbox{yellow}{did he put the limes in the bedroom}\\ Bob: \colorbox{yellow}{yes} \\Alice: where can I find the sweet potatoes\\ Bob: they are in the office or the staircase\\ Alice: where did you see Aiden\\ Bob: I know he didn't went to the staircase\\ Alice: where was he\\ Bob: he said he went to the office\\ Alice: where can I find the carrots\\ Bob: I don't know\end{tabular} &
Did \textbf{Jack} put the \textbf{limes} in the bedroom? &
\begin{tabular}[c]{@{}l@{}}GRICE \\ (Manual)\end{tabular} \\
\midrule

\begin{tabular}[t]{@{}l@{}}"A: Doctor! Doctor! Help me, please!\\ B: Take it easy, please! Sit here, now, what's the matter with you? And can I help you?\\ A: I feel sick. I have a bad stomach-ache, and my head troubles me a lot.\\ B: Well, let me have a check. Open your mouth please. And say ""Ah"".\\ A: Ah! Is that serious, doctor?\\ B: \colorbox{yellow}{Not really. Did you have your supper this evening?}\\ A: \colorbox{yellow}{No, I only had some bananas.}\\ B: \colorbox{yellow}{And the bananas were not quite ripe, right?}\\ A: \colorbox{yellow}{Well, yes, they were a bit green.}\\ B: That explains it.\\ A: I wish I had not eaten them.\\ B: Take this medicine twice a day and I'm sure you'll be fine.\\ A: Thanks, doctor. I'll do as you told me. Good-bye, doctor.\\ B: Bye! And take care."\end{tabular} &
Did \textbf{A} take \textbf{ripe bananas} in the evening? &
\begin{tabular}[c]{@{}l@{}}CICERO \\ (Manual)\end{tabular} \\
\midrule

\begin{tabular}[t]{@{}l@{}}Alice: did you see the turnip?\\ Bob: Yes, Oliver left it in the garage.\\ Alice: did Oliver put the turnip there?\\ Bob: Yes, he placed it there earlier.\\ Alice: \colorbox{yellow}{How about the eggplants?}\\ Bob: \colorbox{yellow}{Sophia took them to the lounge.}\\ Alice: \colorbox{yellow}{Are all the eggplants in the lounge?}\\ Bob: \colorbox{yellow}{Yes, all of them are there.}\\ Alice: Where can I find Sophia?\\ Bob: She said she was in the lounge with the eggplants.\\ Alice: Was Oliver there?\\ Bob: No, he was last seen in the garage.\\ Alice: Where did you see Oliver?\\ Bob: He was in the garage, checking on the turnip.\\ Alice: Did you leave the turnip in the garage?\\ Bob: Yes, I left it there after Oliver brought it.\\ Alice: How many turnips are in the garage?\\ Bob: Just the one that Oliver brought.\\ Alice: Then where did Sophia go after the lounge?\\ Bob: She mentioned she might go to the study next.\end{tabular} &
Did \textbf{Sophia} place all the \textbf{eggplants} at lounge? &
\begin{tabular}[c]{@{}l@{}}GRICE \\ (Synthetic)\end{tabular} \\
\midrule

\begin{tabular}[t]{@{}l@{}}B: \colorbox{yellow}{good afternoon, i'm interested in purchasing a winter jacket.}\\ A: good afternoon! we have a wonderful selection of winter jackets. what size are you looking for?\\ B: i would prefer a medium size. do you have any in blue?\\ A: \colorbox{yellow}{yes, we do have medium-sized jackets available in blue.} would you like to take a look at one?\\ B: yes, that sounds perfect. can you tell me more about the material and warranty?\\ A: certainly! this blue medium jacket is made from high-quality waterproof material and comes with a one-year warranty against any manufacturing defects.\\ B: that sounds reassuring. can i try it on to see how it fits?\\ A: absolutely, the changing room is right over there.\\ B: thank you, i'll check it out.\\ B: \colorbox{yellow}{this fits well and i like the color. i'll take this one.}\\ A: great choice! i'll get it wrapped up for you at the counter.\\ B: thank you for your help.\end{tabular} &
Did the \textbf{buyer} buy the \textbf{blue winter jacket}? &
\begin{tabular}[c]{@{}l@{}}CICERO \\ (Synthetic)\end{tabular} \\
\bottomrule
\end{tabular}%
}
\caption{Examples from \textsc{PragWorld} spanning the two sources of datasets (GRICE, CICERO) and two methods of generation (manual, synthetic).}
\label{tab:examplespragworld}
\end{table*}


\begin{table*}[t]
\centering
\renewcommand{\arraystretch}{1.2}
\begin{adjustbox}{max width=\textwidth}
\begin{tabular}{llcccccccc}
    \toprule
    \multirow{2}{*}{\textbf{Data Type}} & \multirow{2}{*}{\textbf{Accuracy}} & 
    \makecell{\textbf{Variable} \\ \textbf{Swap}} & 
    \makecell{\textbf{Variable} \\ \textbf{Substitution}} & 
    \makecell{\textbf{Quantity} \\ \textbf{Change}} & 
    \makecell{\textbf{Quantifier} \\ \textbf{Change}} & 
    \makecell{\textbf{Logical} \\ \textbf{Connective} \\ \textbf{Change}} & 
    \makecell{\textbf{Not} \\ \textbf{Altered}} & 
    \textbf{Negation} & 
    \makecell{\textbf{Inconsistent} \\ \textbf{Data}} \\
    \midrule
    \multirow{2}{*}{\textbf{Synthetic}} 
    & Yes Acc. & 76.22 & 80.22 & 86.60 & 80.42 & 80.95 & 81.42 & \multicolumn{2}{c}{Not Present} \\
    & No Acc.  & 88.20 & 89.94 & 92.78 & 92.63 & 92.12 & 92.95 & & \\
    \midrule
    \multirow{2}{*}{\textbf{Manual}} 
    & Yes Acc. & 37.50 & 50.88 & 48.48 & 53.85 & 75.00 & 60.42 & 52.38 & 80.00 \\
    & No Acc.  & 92.31 & 97.96 & 87.50 & 88.46 & 100.00 & 92.50 & 100.00 & 90.91 \\
    \bottomrule
\end{tabular}
\end{adjustbox}
\caption{Accuracy distribution across different alteration types for Synthetic and Manual variant of \textsc{PragWorld}. The answers to the questions were generated using \texttt{GPT-3.5-Turbo}. Here, ``Not Altered'' means original or unaltered conversations.}
\label{tab:accuracy_distribution}
\end{table*}

\begin{table}[!htbp]
\centering
\begin{tabular}{llccc}
        \toprule
        & & \textsc{Short} & \textsc{Medium} & \textsc{Long} \\
        \cmidrule(lr){3-3} \cmidrule(lr){4-4} \cmidrule(lr){5-5}
        \multicolumn{1}{c}{\textbf{Data Type}} & \textbf{Accuracy} & 
        \textbf{$<$ 11} & 
        \textbf{11 to 15} &  
        \textbf{$>$ 15} \\
        \midrule
        \textbf{Synthetic} & \textbf{Yes Acc.} & 75.43 & 83.78& 80.43\\
        & \textbf{No Acc.} & 93.93 & 93.82& 90.88\\
        \midrule
        \textbf{Manual} & \textbf{Yes Acc.} & 55.55 & 92.85 & 39.20 \\
        & \textbf{No Acc.} & 93.61 & 94.64 & 93.33 \\
        \bottomrule
\end{tabular}%

\caption{Accuracy distribution (in \%) across different length categories for Synthetic and Manual variants of the \textsc{PragWorld} dataset. The answers to the questions were generated using \texttt{GPT-3.5-Turbo}.}
\label{tab:accuracy_distribution_across_length}
\end{table}

\paragraph{Label distribution}
\label{app:sub-data-dist}
\Cref{fig:yes_no_acc_finegrained_distribution} shows the distribution of questions with gold answer as "Yes" or "No" across the manual and synthetic variants of the \textsc{PragWorld} dataset. The figures shows that the number of questions with gold answer as "Yes" are comparable to those with gold answer as "No" in all the variants.

We report accuracy distribution across different alteration types for the synthetic and manual splits of \datasetname{} in \Cref{tab:accuracy_distribution}, while we report performance across different conversation length categories (wrt. number of utterances) in \Cref{tab:accuracy_distribution_across_length}. \Cref{tab:additionalresults_globalresultstable} shows the performance of models with larger parameters (Llama-3.1-70B-Instruct, Llama-3.3-70B-Instruct, Qwen2.5-32B-Instruct, and Qwen2.5-72B-Instruct) on both the manual and synthetic variant of \datasetname{}.

\begin{table*}[t]
    \centering
    \renewcommand{\arraystretch}{1.5}
    \begin{tabular}{|p{3cm}|p{10cm}|}
        \hline
        \textbf{Dataset Name} & \textbf{Conversation Example} \\
        \hline
        GRICE & Person 1: Where did Emma go? \\ 
                   & Person 2: She said she was in the cellar. \\ 
                   & Person 1: Did you leave the oranges in the garage? \\
                   & Person 2: I left the watermelons there. \\
                   & Person 1: Did Noah leave them there? \\
                   & Person 2: He didn't. \\
                   & Person 1: Are some of the strawberries in the backyard? \\
                   & Person 2: Two are there. \\
        \hline
        CICERO & A: Excuse me. I'd like to find out about flights to New York. \\
                   & B: Well, let's see. One just left about five minutes ago; and there's another one at ten. \\
                   & A: What time is it, please? \\
                   & B: It's five to eight. \\
                   & A: So the plane leaves in about two hours. \\
                   & B: That's right. Have you bought your ticket? \\
                   & A: No, I haven't. Can I buy one here? \\
                   & B: I'm afraid you can't. You'd better go to the booking office. \\
        \hline
    \end{tabular}
    \caption{Examples of seed conversations from the GRICE and CICERO datasets.}
    \label{tab:seedconversations}
\end{table*}

\label{app:stats-question-type}
\begin{table*}[t]
    \centering
    \renewcommand{\arraystretch}{1.2}
    \begin{adjustbox}{max width=\textwidth}
    \begin{tabular}{llccccccc}
        \toprule
        \multirow{2}{*}{\textbf{Data Type}} & \multirow{2}{*}{\textbf{Accuracy}} & 
        \makecell{\textbf{Quantity} \\ \textbf{Ques}} & 
        \makecell{\textbf{Univ. Quant.} \\ \textbf{Ques}} & 
        \makecell{\textbf{Exis. Quant.} \\ \textbf{Ques}} & 
        \makecell{\textbf{OnlyEntity} \\ \textbf{Ques}} & 
        \makecell{\textbf{Sure} \\ \textbf{Ques}} & 
        \makecell{\textbf{Know} \\ \textbf{Ques}} & 
        \makecell{\textbf{Other} \\ \textbf{Ques}} \\
        \midrule
        & \textbf{Yes Acc.} & 96/134 & 54/126 & 343/357 & \multicolumn{4}{c}{Not Present} \\
        \textbf{Synthetic} & \textbf{No Acc.} & 220/257 & 263/268 & 14/37 & & & & \\
        \midrule
        & \textbf{Yes Acc.} & 3/6 & 0/1 & 15/17 & 41/66 & \textbf{0/57} & \textbf{24/49} & 63/81 \\
        \textbf{Manual} & \textbf{No Acc.} & 8/8 & 5/5 & 5/5 & 53/56 & 33/33 & 29/34 & 76/82 \\
        \bottomrule
    \end{tabular}
    \end{adjustbox}
    \caption{Accuracy distribution in fractional notation across different Question types for Synthetic (generated from GRICE) and Manual variants of the \textsc{PragWorld} dataset. The answers to the questions were generated using \texttt{GPT-3.5-Turbo}, which struggles in the "Sure-Ques", and "Know-Ques" categories (highlighted in Bold). Due to the complexity of conversations in the CICERO version of the synthetic \textsc{PragWorld} dataset, categorizing questions in this variant is challenging. Consequently, we have included results only for the GRICE synthetic variant of the \textsc{PragWorld} dataset. Here, \textbf{Univ. Quant. Ques}, and \textbf{Exis. Quant. Ques} means Universal Quantifier and Existential Quantifier questions respectively.}
    \label{tab:question_distribution-fractional-fromat}
\end{table*}


\section{Hardware and Experimental Details}
\begin{table*}[t]
\resizebox{\linewidth}{!}{%
\begin{tabular}{lccccc}
        \hline
        \textbf{Model name} & \textbf{Parameters} & \textbf{Context Size} & \textbf{Decoding Method} & \textbf{Temperature} & \textbf{Top-p} \\ \hline
        deepseek-ai/DeepSeek-Coder-V2-Lite-Instruct                   & 16 B (2.4 B active)      & 128K   & Greedy Decoding  & 0.0  & 1.0 \\ \hline
        openai/gpt-3.5-turbo                    & -      & 16K   & Greedy Decoding  & 0.0  & 1.0 \\ \hline
        microsoft/Phi-3-mini-4k-instruct        & 3.8 B  & 4K    & Greedy Decoding  & 0.0  & 1.0 \\
        microsoft/Phi-3.5-mini-instruct         & 3.8 B  & 128K  & Greedy Decoding   & 0.0  & 1.0 \\ \hline
        meta-llama/Llama-3.2-1B-Instruct        & 1 B    & 128K  & Greedy Decoding  & 0.0  & 1.0 \\
        meta-llama/Llama-3.2-3B-Instruct        & 3 B    & 128K  & Greedy Decoding  & 0.0  & 1.0 \\
        meta-llama/Llama-3.1-8B-Instruct        & 8 B    & 128K  & Greedy Decoding  & 0.0  & 1.0 \\
        meta-llama/Llama-3.1-70B-Instruct        & 70 B    & 128K  & Greedy Decoding  & 0.0  & 1.0 \\
        meta-llama/Llama-3.3-70B-Instruct        & 70 B    & 128K  & Greedy Decoding  & 0.0  & 1.0 \\\hline
        Qwen/Qwen2.5-0.5B-Instruct              & 0.5 B  & 128K  & Greedy Decoding  & 0.0  & 1.0 \\
        Qwen/Qwen2.5-1.5B-Instruct              & 1.5 B  & 128K  & Greedy Decoding  & 0.0  & 1.0 \\
        Qwen/Qwen2.5-7B-Instruct                & 7 B    & 128K  & Greedy Decoding  & 0.0  & 1.0 \\
        Qwen/Qwen2.5-32B-Instruct                & 32 B    & 128K  & Greedy Decoding  & 0.0  & 1.0 \\
        Qwen/Qwen2.5-72B-Instruct                & 72 B    & 131K  & Greedy Decoding  & 0.0  & 1.0 \\ \hline
\end{tabular}%
}   
\caption{Baseline models used in our experiments, including their model names, parameter sizes, context lengths, and decoding parameters.}
\label{tab:baseline_models}
\end{table*}

All the open-source models listed in Table \ref{tab:baseline_models} were fine-tuned using two NVIDIA A40 GPUs, each with $48$ GB of CUDA memory.
The fine-tuning process was conducted with the LLaMA-Factory library \cite{zheng2024llamafactory}, using the LoRA rank of $8$, a learning rate of 1e-4, three epochs, and a warm-up ratio of $0.1$. Additionally, LoRA adapters were applied to all relevant layers, including the attention projection layers (q\_proj, k\_proj, v\_proj, o\_proj), the feedforward network layers (gate\_proj, up\_proj, down\_proj), and other model components such as embed\_tokens and lm\_head. For \texttt{GPT-3.5}, we used a temperature of $0.0$, top\_p = $1.0$, and max\_tokens = $1024$ while generating the responses. 

\section{Additional Results on \datasetname{}}
Table \ref{tab:additionalresults_globalresultstable} shows the performance of the models Llama-3.1-70B-Instruct, Llama-3.3-70B-Instruct, Qwen2.5-32B-Instruct, and Qwen2.5-72B-Instruct models on the manual variant of the \datasetname{}. These models have a larger number of parameters as compared to the models shown in the main paper.

\section{Task Prompts}
\label{subsec:taskprompts}

When prompting LMs to solve \datasetname{}, we first present the model with the conversation, and then the question. This setup necessitates the model memorize all information present in the context prior to answering. We detail the prompts used to produce answers in \Cref{tab:inference_prompts}.

\begin{table*}[t]
\centering
\resizebox{\textwidth}{!}{%
\begin{tabular}{p{0.96\linewidth}}
\toprule
\textbf{Prompt 1} \\
\midrule
Read this conversation between Bob and Alice: \\
\texttt{\{context\}} \\

Now based on your understanding of the conversation, answer the question below: \\
\texttt{\{question\}} \\

Answer this with only a (yes) or (no) in the first line and then explain your answer from the next line onwards. \\
Your answer: \\
( \\
\midrule

\textbf{Prompt 2} \\
\midrule
Read this conversation between the two speakers: \\
\texttt{\{context\}} \\

Now, based on your understanding of the conversation, answer the question below: \\
\texttt{\{question\}} \\

The answer should only be a label, i.e,. either yes or no. \\
Label: \\
\bottomrule
\end{tabular}%
}
\caption{The table shows two prompts: \textbf{Prompt 1}: This prompt is given to both the open-source and closed-source models while generating answers to the questions. \textbf{Prompt 2}: This prompt is given to the open-source models during fine-tuning and further inferencing.}
\label{tab:inference_prompts}
\end{table*}

\section{Prompts Used to Create the Synthetic Dataset}
\label{subsec:syntheticprompts}
We designed prompts to be provided as input to the \texttt{GPT-4} model, along with a seed or original conversation, to generate synthetic conversations. In all the prompts, \verb|[[$$SEED_CONVERSATION$$]]| is used as a place holder for the seed conversation taken either from the GRICE or CICERO datasets. While generating the conversations using \texttt{GPT-4}, we used a temperature of 0.7, set the value of top\_p value as 0.95, and max\_tokens value to 800 to allow for sufficiently detailed outputs.
We show the prompts used in \Cref{tab:synth-prompt-1,tab:synth-prompt-2,tab:synth-prompt-3}.

\begin{table*}[t]
\centering
\begin{tabularx}{\textwidth}{|X|}
\hline
\textbf{Prompt used to create the synthetic version from the GRICE conversations.} \\
\hline
Given below is a seed conversation where the first speaker enquires to the second speaker about some \textless objects\textgreater{} and \textless agents\textgreater{}. \\

\textbf{Enquiry about the \textless objects\textgreater{}}: \\
\textbf{A) Location of the \textless objects\textgreater{}}: \\
1. did you see the \textless object\textgreater{}? ("you" implies the second speaker) \\
2. what about the \textless object\textgreater{}? \\
3. where can i get the \textless objects\textgreater{}? \\
4. where can i find the \textless objects\textgreater{}? \\
5. how about the \textless objects\textgreater{}? \\

\textbf{B) Quantity of those \textless objects\textgreater{} at a particular location.} \\
1. how many \textless objects\textgreater{} are in the \textless location\textgreater{}? \\
2. Are all \textless objects\textgreater{} there? ("there" implies a \textless location\textgreater{} in the previous utterances.) \\
3. are some of them in the \textless location\textgreater{}? ("some" implies an \textless object\textgreater{} in the previous utterances.) \\

\textbf{Enquiry about the \textless agents\textgreater{}}: \\
\textbf{A) Location of the \textless agents\textgreater{}} \\
1. were you there? ("you" implies the second speaker.) \\
2. then where? \\
3. was \textless agent\textgreater{} there? \\
4. where did you see \textless agent\textgreater{}? ("you" implies the second speaker.) \\
5. where was he? ("he" implies an \textless agent\textgreater{} mentioned earlier.) \\

\textbf{B) \textless agents\textgreater{} who put those \textless objects\textgreater{} at a particular location.} \\
1. did \textless agent\textgreater{} place the \textless objects\textgreater{} in the \textless location\textgreater{}? \\
2. did \textless agent\textgreater{} put the \textless objects\textgreater{} there? (”there” implies a \textless location\textgreater{} mentioned in
the previous utterances.)\\
3. did you put them there? (”you”, ”them”, and ”there” imply second speaker, \textless object\textgreater{}, and \textless location\textgreater{} respectively mentioned in the previous utterances.)\\
4. did he leave the \textless objects\textgreater{} in the \textless location\textgreater{}?  (”he” implies an \textless agent\textgreater{} mentioned
in the previous utterances.)\\

\textbf{You have to perform the following tasks:} \\
    \textbf{Task 1:} Identify all the \textless agents\textgreater{}, \textless objects\textgreater{}, and \textless locations\textgreater{} from the seed conversation excluding the speakers. \\
    \textbf{Task 2:} Select two \textless agents\textgreater{} (say A1, and A2), \textless objects\textgreater{} (say O1, and O2), and \textless locations\textgreater{} (say L1, and L2) identified in Task 1 at random and create two triplets i.e., (A1, O1, L1), and (A2, O2, L2). \\
    \textbf{Task 3:} Create a conversation using the \textless agents\textgreater{}, \textless objects\textgreater{}, and \textless locations\textgreater{} identified in Task 1 with special focus on the triplets (A1, O1, L1), and (A2, O2, L2) created in Task 2. You can take help from the aforementioned examples regarding enquiry about the \textless objects\textgreater{} and \textless agents\textgreater{} to create the conversations. \\
    
    \textbf{Note:} It is important that the structure of conversation should be such that it requires creating a world model with respect to both the speakers which maintains the states while iterating through all 10 pairs of utterances for answering the question. \\
    
    \textbf{Show the Output of all the tasks like:} \\
    \textbf{Task 1 Output:} \\
    \textless agents\textgreater{}: \\
    \textless objects\textgreater{}: \\
    \textless locations\textgreater{}: \\
    
    \textbf{Task 2 Output:} \\
    Selected Agents: \\
    Selected Objects: \\
    Selected Locations: \\
    First Triplet (A1, O1, L1): \\
    Second Triplet (A2, O2, L2): \\
    
    \textbf{Task 3 Output:} \\
    Generated Conversation: \\
    
    \textbf{Seed Conversation:} \\
    \verb|[[$$SEED_CONVERSATION$$]]| \\
\hline
\end{tabularx}
\caption{Prompt given to GPT-4 to generate synthetic conversations from GRICE dataset seed samples.}
\label{tab:synth-prompt-1}
\end{table*}

\begin{table*}[t]
\centering
\begin{tabular}{|p{0.96\linewidth}|}
\hline
\textbf{Prompt used to create the synthetic version from the CICERO Seller-Buyer Conversations} \\
\hline
Given below is a seed conversation where the first speaker enquires to the second speaker about some \textless items\textgreater{} and its \textless properties\textgreater{} like color, weight, quantity, quality, variation, cost, warranty, region of availability, etc. \\
    
    \textbf{You have to perform the following tasks:} \\
    \textbf{Task 1:} Identify all the \textless agents\textgreater{}, \textless items\textgreater{}, and \textless properties\textgreater{} from the seed conversation excluding the speakers. \\
    \textbf{Task 2:} Select two \textless agents\textgreater{} (say A1, and A2), \textless items\textgreater{} (say I1, and I2), and \textless properties\textgreater{} (say P1, and P2) identified in Task 1 at random, and create two triplets i.e., (A1, I1, P1), and (A2, I2, P2). \\
    \textbf{NOTE:} These two triplets i.e., (A1, I1, P1), and (A2, I2, P2) should not have the same entity. Generate a new entity if required. \\
    \textbf{Task 3:} Create a new conversation using the \textless agents\textgreater{}, \textless items\textgreater{}, and \textless properties\textgreater{} identified in Task 1 with special focus on the triplets (A1, I1, P1), and (A2, I2, P2) created in Task 2. The new conversation about both the triplets should include the structure of a Seller-Buyer conversation. In this setup, the first speaker takes on the role of the buyer/customer, while the second speaker acts as the seller. \\
    
    1) The buyer inquires about certain \textless items\textgreater{} from the seller. \\
    2) The seller describes or displays those \textless items\textgreater{} to the buyer. \\
    3) The buyer asks in detail about the properties of those \textless items\textgreater{} from the seller. The \textless properties\textgreater{} can be color, weight, quantity, quality, variation, cost, warranty, region of availability, etc. \\
    4) The seller provides the requested details and presents all available options to the buyer. \\
    5) In some cases, the buyer asks for the seller’s recommendation on which \textless item\textgreater{} to choose, but this does not occur in every conversation. \\
    6) Finally, the buyer decides whether to purchase some \textless items\textgreater{} from the seller. In some cases, the buyer chooses not to buy any \textless items\textgreater{} due to different preferences regarding the \textless properties\textgreater{}, which the seller’s current inventory does not meet. \\
    
    \textbf{Note:} It is important that the structure of conversation should be such that it requires creating a world model with respect to both the speakers which maintains the states. \\
    
    \textbf{Show the Output of all the tasks like:} \\
    \textbf{Task 1 Output:} \\
    \textless agents\textgreater{}: \\
    \textless items\textgreater{}: \\
    \textless properties\textgreater{}: \\
    \textbf{Task 2 Output:} \\
    Selected Agents: \\
    Selected Items: \\
    Selected Properties: \\
    First Triplet (A1, I1, P1): \\
    Second Triplet (A2, I2, P2): \\
    
    \textbf{Task 3 Output:} \\
    Generated Conversation: \\
    
    \textbf{Seed Conversation:} \\
    \verb|[[$$SEED_CONVERSATION$$]]| \\
\hline
\end{tabular}
\caption{A prompt for \texttt{GPT-4} to create a synthetic conversation from a given ``Seller-Buyer'' type seed conversation from the CICERO dataset.}
\label{tab:synth-prompt-2}
\end{table*}

\begin{table*}[t]
\centering
\begin{tabular}{|p{0.96\linewidth}|}
\hline
\textbf{Prompt used to create the synthetic version from the CICERO Doctor-Patient Conversations} \\
\hline
 Given below is a seed conversation where the first speaker enquires to the second speaker about topics like Doctor-patient appointment scheduling, Health Advice, or Medical Consultation. The conversation involves some \textless agents\textgreater{}, \textless medical entities\textgreater{}, and \textless locations\textgreater{}. \\
    
    \textbf{You have to perform the following tasks:} \\
    \textbf{Task 1:} Identify all the \textless agents\textgreater{}, \textless medical entities\textgreater{}, and \textless locations\textgreater{} from the seed conversation excluding the speakers. \\
    \textbf{Task 2:} Select two \textless agents\textgreater{} (say A1, and A2), \textless medical entities\textgreater{} (say M1, and M2), and \textless locations\textgreater{} (say L1, and L2) identified in Task 1 at random, and create two triplets i.e., (A1, M1, L1), and (A2, M2, L2). \\
    \textbf{NOTE:} These two triplets i.e., (A1, M1, L1), and (A2, M2, L2) should not have the same entity. Generate a new entity if required. \\
    \textbf{Task 3:} Create a new conversation using the \textless agents\textgreater{}, \textless medical entities\textgreater{}, and \textless locations\textgreater{} identified in Task 1 with special focus on the triplets (A1, M1, L1), and (A2, M2, L2) created in Task 2. The new conversation about both the triplets should be created around all of the five following questions: \\
    
    1) \textbf{Causes:} What \textless medical entities\textgreater{} caused \textless agent\textgreater{} to go to the \textless location\textgreater{}? \\
    2) \textbf{Subsequent Events:} What \textless agent\textgreater{} did regarding the \textless medical entities\textgreater{} after arriving at the \textless location\textgreater{}? \\
    3) \textbf{Prerequisite:} What happened before with the \textless agent\textgreater{} regarding \textless medical entities\textgreater{} that made \textless agent\textgreater{} visit the \textless location\textgreater{}? \\
    4) \textbf{Motivation:} What motivated the \textless agent\textgreater{} to visit the \textless location\textgreater{} regarding the \textless medical entities\textgreater{}? \\
    5) \textbf{Reaction:} How \textless agent\textgreater{} reacted regarding the \textless medical entities\textgreater{} after reaching the \textless location\textgreater{}? \\
    
    In the aforementioned questions, \textless agent\textgreater{}, \textless medical entities\textgreater{}, and \textless location\textgreater{} are the placeholders for the triplets (A1, M1, L1), and (A2, M2, L2). \\
    
    \textbf{Note:} It is important that the structure of conversation should be such that it requires creating a world model with respect to both the speakers which maintains the states. \\
    
    \textbf{Show the Output of all the tasks like:} \\
    \textbf{Task 1 Output:} \\
    \textless agents\textgreater{}: \\
    \textless medical entities\textgreater{}: \\
    \textless locations\textgreater{}: \\
    
    \textbf{Task 2 Output:} \\
    Selected Agents: \\
    Selected Medical Entities: \\
    Selected Locations: \\
    First Triplet (A1, M1, L1): \\
    Second Triplet (A2, M2, L2): \\
    
    \textbf{Task 3 Output:} \\
    Generated Conversation: \\
    
    \textbf{Seed Conversation:} \\
    \verb|[[$$SEED_CONVERSATION$$]]| \\
\hline
\end{tabular}
\caption{A prompt for \texttt{GPT-4} to create a synthetic conversation from a given ``Doctor-Patient'' type seed conversation from the CICERO dataset.}
\label{tab:synth-prompt-3}
\end{table*}

\section{Dual-perspective Interpretability Framework}

\subsection{Notation and Preliminaries}

Let $(x + q, y^{\text{gold}}) \in \mathcal{D}$ denote an original conversation prompt and its gold label, and let $(\hat{x}+\hat{q}, \hat{y}^{\text{gold}}) \in \mathcal{D}$ denote its corresponding altered variation prompt and gold label. Here, $x$ and $q$ denote the original conversation and question, while $\hat{x}$ and $\hat{q}$ represent the altered conversation and question, respectively. Let $y$ and $\hat{y}$ denote the predicted labels for the original prompt $x + q$ and its altered variation $\hat{x} + \hat{q}$, respectively.  
 
Here, $y^{\text{gold}}, y, \hat{y}^{\text{gold}}, \hat{y} \in \{\texttt{yes}, \texttt{no}\}$. For each transformer layer $\ell=1,\dots,L$ we will consider two interventions:
\begin{itemize}
  \item \textbf{Direct Effect Patching:} Here, we measure the Direct Effect by injecting layer~$\ell$’s residual activation computed on the altered prompt into the forward pass of the original prompt. This patch isolates how much that single layer’s update shifts the final probability.
  \item \textbf{MLP zero‐out Ablation:} Here, we disable the feed‐forward submodule at layer~$\ell$ at a given layer by zeroing out its output.
\end{itemize}

\subsection{Direct Effect Patching}
The first step in this intervention is to filter the dataset $\mathcal{D}$ as follows: For a given original prompt $(x, y^{\text{gold}})$ and its altered variant $(\hat{x}, \hat{y}^{\text{gold}})$, we select this altered variant only if $\bigl(\hat{y} =  \hat{y}^{\text{gold}}\bigr) \land \bigl(\hat{y} \neq y\bigr)$, i.e., the prediction of the model is correct for the altered variant and it predicts an opposite label for the original variant. This is done so that we can observe how the confidence of the model shifts to predict the opposite label when we replace the residual stream at layer~$\ell$ in the original conversation run with that from the altered conversation run. We call the filtered dataset as $\mathcal{D}^{'}$. Let $P\bigl(\hat{y}^{\text{gold}}\mid x+q\bigr)$ be the probability that the model predicts the gold label for the altered prompt when the input is the original conversation prompt. We call this probability \textbf{Original Run (OR)}. Similarly, Let $P\bigl(\hat{y}^{\text{gold}}\mid \hat{x}+\hat{q}\bigr)$ be the probability that the model predicts the gold label for the altered prompt when the input is the altered conversation prompt. We call this probability \textbf{Altered Run (AR)}. Let $R^{P}_l = \texttt{Residual}_l(P)$ be the layer~$\ell$ residual stream activations produced by a normal forward pass with $P$ as the input prompt. We now define \textbf{Direct Effect} at layer~$\ell$ as the change in probability obtained by patching the altered residuals in the original run: 
\begin{align*}
DE(R^{\hat{x}}_\ell \rightarrow R^{x}_\ell) 
&= P\bigl(\hat{y}^{\text{gold}} \mid x+q;\, \texttt{patch}_\ell(R^{\hat{x}}_\ell)\bigr) \\
&\quad - P\bigl(\hat{y}^{\text{gold}} \mid x+q \bigr)
\end{align*}
where, $\texttt{patch}_\ell()$ is the patching operator that replaces layer~$\ell$ residuals with $R^{\hat{x}}_\ell$, and $R^{\hat{x}}_\ell \rightarrow R^{x}_\ell$ indicates that we patch from the altered run into the original run. This \textit{direct effect} highlights layers where new altered input triggers large confidence shifts. 

\paragraph{Difference from the previous approach.} We measure the Direct Effect by injecting layer~$\ell$’s residual activation computed on the altered prompt into the forward pass of the original prompt. This patch isolates how much the single-layer update changes the final probability. Although we follow \citet{joshi-etal-2025-towards}, the ready availability of the paired original and its altered variants in our dataset makes the analysis more feasible and straightforward.

\subsection{MLP Zero‐out Ablation}
Let $P^{(\mathrm{MLP}_\ell \,\rightarrow \,0)}\bigl(y \mid x+q\bigr)$ denote the model’s predicted probability of label \(y\) on input \(x + q\) when the MLP submodule at layer \(\ell\) has been zeroed out.  The predicted label after ablating the MLP submodule at layer~$\ell$ is given by:
\[
\hat{y}
\;=\;
\arg\max_{\text{y}\in\{\texttt{yes},\texttt{no}\}}
\;P^{(\mathrm{MLP}_\ell \,\rightarrow \,0)}\bigl(y \mid x+q\bigr)
\] where $(x+q, y) \in \mathcal{D}$. The accuracy over the entire dataset $\mathcal{D}$ after zeroing out MLP submodule at layer~$\ell$ is given by:
\[
A_\ell
\;=\;
\frac{1}{|\mathcal{D}|}
\sum_{i=1}^{|\mathcal{D}|}
\mathbf{1}\!\Bigl[\hat{y}_{i} = y_{i}\Bigr]
\]
We compare \(A_\ell\) to the baseline (no‑ablation) accuracy as:
\[
A_0
\;=\;
\frac{1}{|\mathcal{D}|}
\sum_{i=1}^{|\mathcal{D}|}
\mathbf{1}\!\Bigl[\arg\max_{\text{y}_i\in\{\texttt{yes},\texttt{no}\}}
\;P\bigl(y_i \mid x_i+q_i\bigr) = y_{i}
\Bigr]
\]
We classify layer \(\ell\) as \textbf{useful} if $A_\ell < A_0$, and \textbf{harmful} if $A_\ell > A_0$. 

\paragraph{Difference from the previous approach.} We differ from \citet{geva2023dissectingrecallfactualassociations} in that we perform MLP zero-out ablation across all aligned token positions for all layers of a given model, and measure the resulting change in accuracy. In contrast, they zero-out MLPs at a single ``subject'' token position for $10$ consecutive layers.

\begin{figure}[h]
    \centering
    \includegraphics[width=1.0\linewidth]{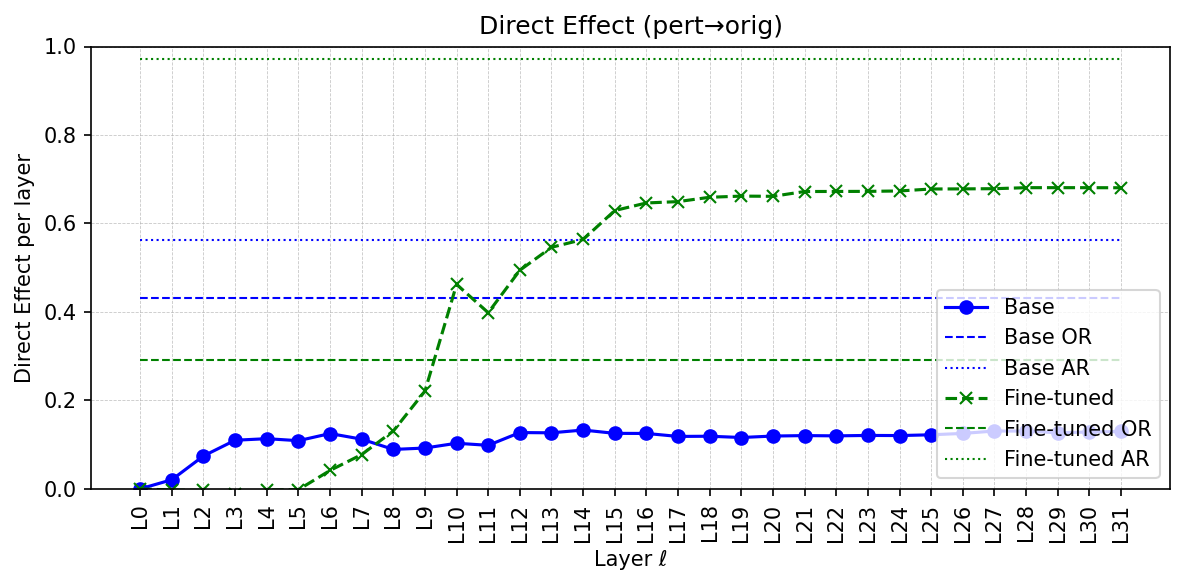}  
    \caption{Comparison of the Direct Effect Probing on both the base and fine-tuned version of the \texttt{Phi-3.5-mini-instruct} model for a given example where the base version didn't make the correct prediction on the original conversation. Original Run (OR) is the probability of a model to predict the gold label for the altered conversation when the input is the original conversation. Altered Run (AR) is the probability of a model to predict the gold label for the altered conversation when the input is the altered conversation. As we have patched the altered residuals into the original run, we expect the model's confidence to shift more towards AR, and far away from OR.}
    \label{fig:DE_Phi_incorrect_raw_ftuned}
\end{figure}

\paragraph{Observations from Direct Effect Patching on an \textit{incorrect} example.}
Figure \ref{fig:DE_Phi_incorrect_raw_ftuned} shows the Direct effect for an example where the base version is correct only on the altered conversation, but not on the original conversation, while the fine-tuned model is correct on both the original and altered conversation. Here, also, we see that the confidence of the fine-tuned model increases drastically after layer 5, while for the base model, the confidence of the model didn't go above 0.2 for all the layers. Here, also, the fine-tuned version shows a low \textbf{OR} and high \textbf{AR} confidence as compared to the base version.

\section{Targeted layer Regularization}

Guided by the observations from the \textit{Direct Effect Patching} and \textit{MLP Zero-out Ablation} experiments,  we define two regularization techniques targeting the \textit{useful} and the \textit{harmful} layers. We first define the loss function used in our targeted regularization approach as: 
\[
\mathcal{L}
\;=\;
\mathcal{L}_{\mathrm{CE}}
\;+\;
\alpha\,\mathcal{L}_{\mathrm{amplify}}
\;+\;
\beta\,\mathcal{L}_{\mathrm{suppress}}
\]
Here, $\mathcal{L}_{\mathrm{CE}}$ is the usual cross‐entropy loss on the model’s predicted distribution. For each layer identified as \textit{useful}, we attach a small two‐layer classification head on top of that layer’s MLP output and compute its cross‐entropy against the true label; \(\mathcal{L}_{\mathrm{amplify}}\) is the mean of these probe losses across all useful layers.
  \(\mathcal{L}_{\mathrm{suppress}}\) is the mean squared activation of each \textit{harmful} MLP sublayer, averaged first over all tokens and examples and then across all harmful layers.
Mathematically, we express these loss functions as shown below: \\
\textbf{1. Useful Layer Amplification:}  
    Let $U=\{\ell: A_\ell<A_0\}$ be the set of the useful layers. For each useful layer $\ell$, we attach a small classifier 
$f_\ell : \mathbb{R}^D \rightarrow \mathbb{R}^2$ 
to the final MLP output $h^{i}_\ell \in \mathbb{R}^D$ for instance $i$ in the dataset $\mathcal{D}$, and define the \textit{Amplify Loss} as:

\[
\mathcal{L}_{\text{amplify}} = 
\frac{1}{|U|} \sum_{\ell \in U} \;
\frac{1}{|\mathcal{D}|} \sum_{i=1}^{|\mathcal{D}|} 
\left[ 
    \texttt{CE}\bigl( f_\ell(h_l^i), y_i \bigr)
\right]
\]

\textbf{2. Harmful Layer Suppression:}  
    Let $H=\{\ell : A_\ell>A_0\}$ be the set of the harmful layers, and let $h_\ell^{(i)}\in\mathbb{R}^{D}$ denotes the MLP’s output (the residual vector just after the feed‐forward submodule) at layer~$\ell$.  We add an L2 penalty on these activations. We define the \textit{Supress Loss} as the mean squared norm of those activations, averaged over the entire dataset as shown below:
\[
\mathcal{L}_{\mathrm{suppress}}
\;=\;
\frac{1}{|H|}\sum_{\ell\in H}
\frac{1}{|\mathcal{D}|}\sum_{i=1}^{|\mathcal{D}|}
\bigl\lVert h_{\ell}^{(i)}\bigr\rVert_{2}^{2}
\]

\subsection{Result of Targeted layer Regularization}
\begin{figure}[h]
    \centering
    \includegraphics[width=1.0\linewidth]{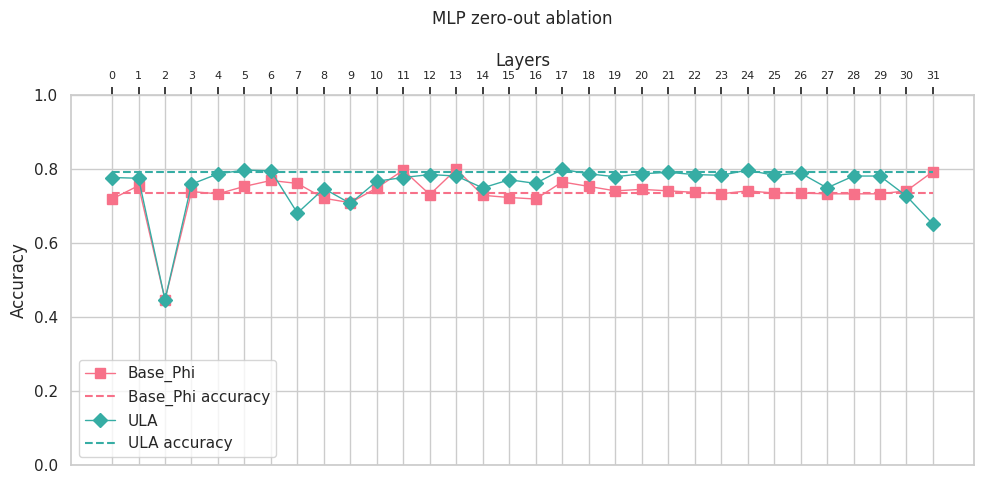}  
    \caption{Results of MLP zero-out ablation on \texttt{Phi-3.5-mini-ins} finetuned using ULA. Base accuracy denotes the base model accuracy on the entire dataset, and ULA accuracy is the accuracy after finetuning using ULA regularization technique.}
    \label{fig:mla_phi_raw_ula_supp}
\end{figure}

\begin{figure}[h]
    \centering
    \includegraphics[width=1.0\linewidth]{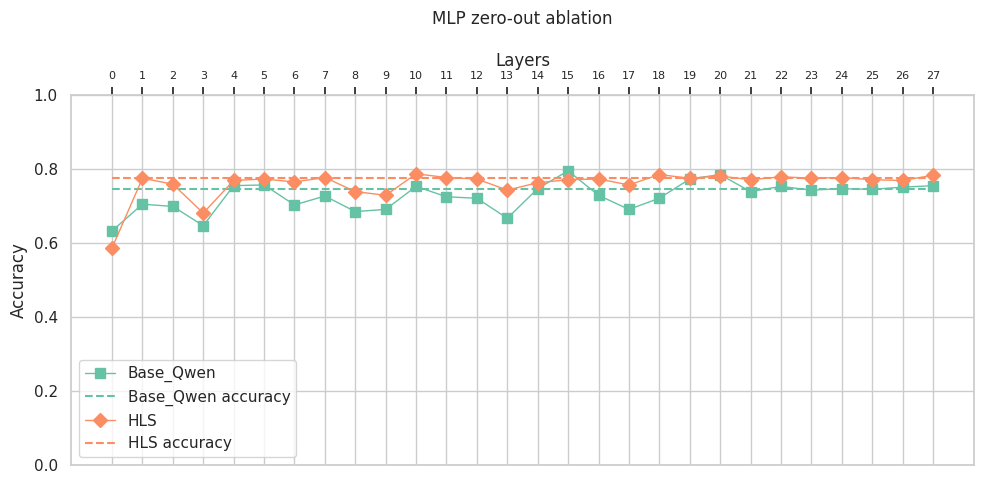}  
    \caption{Results of MLP zero-out ablation on \texttt{Qwen2.5-7B-Ins} finetuned using HLS. Base accuracy denotes the base model accuracy on the entire dataset, and HLS accuracy is the accuracy after finetuning using HLS regularization technique.}
    \label{fig:mla_qwen_raw_hls_supp}
\end{figure}

Figure \ref{fig:mla_phi_raw_ula_supp} shows the effect of \textit{Useful Layer Amplification (ULA)} (\textit{blue line}) regularization technique (for useful layers 2 and 9 visible in Base\_Phi, with \textit{useful weight} of $1e-3$) on \texttt{Phi-3.5-mini-instruct} after MLP zero-out ablation, where it reduced the effect of these harmful layers, as all the layers show accuracy less than the \textit{ULA accuracy}. Similarly, Figure \ref{fig:mla_qwen_raw_hls_supp} shows the effect of \textit{Harmful Layer Suppression (HLS)} (\textit{orange line}) regularization technique (for harmful layers 5, 15, 19, and 20 visible in Base\_Qwen, with \textit{harmful weight} of $1e-3$) on \texttt{Qwen2.5-7B-Instruct} after MLP zero-out ablation, where it reduced the effect of these harmful layers, as all the layers show accuracy less than the \textit{HLS accuracy}. 

\section{Comparison with other datasets}

Table \ref{tab:datasetcmparison} compares our \datasetname{} dataset with existing multi-turn or single-turn conversational, entity-tracking, and pragmatic datasets.
\begin{table*}[!htb]
\resizebox{\textwidth}{!}{%
\begin{tabular}{@{}ccccccc@{}}
\toprule
\textbf{Dataset}                                                                 & \textbf{Multi-Turn}                                      & \begin{tabular}[c]{@{}c@{}}Needs Entity\\Tracking \end{tabular} & \begin{tabular}[c]{@{}c@{}}Needs\\Anaphora \end{tabular}   & \textbf{Natural?}                                            & \textbf{Implicature} & \textbf{Alterations}                                        \\ \midrule
GRICE \citep{zheng-etal-2021-grice}                                                      & \cmark                                                   & \cmark                                                   & \cmark                                                   & \xmark                                                       & \cmark  & \xmark   \\
CICERO  \citep{ghosal-etal-2022-cicero}                                                      & \cmark                                                   & \begin{tabular}[c]{@{}c@{}}\ding{52}*\\ \end{tabular}  & \begin{tabular}[c]{@{}c@{}}\ding{52}*\\ \end{tabular}  & \begin{tabular}[c]{@{}c@{}}\ding{52}*\\ \end{tabular} & \begin{tabular}[c]{@{}c@{}}\ding{52}*\\ \end{tabular}  & \xmark \\
Entity Tracking \cite{Kim2024CodePI}                                          & \xmark                                                   & \cmark                                                   & \xmark                                                   & \xmark                                                       & \xmark                                                        & \xmark\\
MuTual   \citep{cui-etal-2020-mutual}        & \cmark                                                   & \cmark                                                   & \cmark                                                   & \cmark                                                       & \xmark                  & \xmark                                      \\
DROP   \citep{dua-etal-2019-drop}          & \xmark                                                   & \cmark                                                   & \xmark                                                   & \xmark                                                       & \xmark                                                        & \xmark\\
DREAM   \citep{sun-etal-2019-dream}          & \cmark                                                   & \cmark                                                   & \cmark                                                   & \cmark                                                       & \xmark & \xmark                                                       \\
CoQA  \citep{reddy-etal-2019-coqa}            & \cmark                                                   & \xmark                                                   & \cmark                                                   & \cmark                                                       & \xmark   & \xmark                                                     \\
Cosmos QA  \citep{huang-etal-2019-cosmos}      & \xmark                                                   & \cmark                                                   & \cmark                                                   & \cmark                                                       & \xmark   & \xmark                                                     \\
DiPlomat  \citep{10.5555/3666122.3668152}       & \cmark                                                   & \cmark                                                   & \cmark                                                   & \cmark                                                       & \cmark  & \xmark                                                      \\ \midrule
{\color[HTML]{333333} \textbf{\textsc{PragWorld} (ours)}}  & \begin{tabular}[c]{@{}c@{}}\cmark\\ \end{tabular} & \begin{tabular}[c]{@{}c@{}}\ding{52}*\\ \end{tabular} & \begin{tabular}[c]{@{}c@{}}\ding{52}*\\ \end{tabular} & \begin{tabular}[c]{@{}c@{}}\ding{52}*\\ \end{tabular}     & \begin{tabular}[c]{@{}c@{}}\ding{52}*\\ \end{tabular}   & \cmark   \\ \bottomrule
\end{tabular}%
}
\caption{We compare existing multi-turn or single-turn conversational, entity-tracking, and pragmatic datasets that share many features with our work. Our work provides a balance over many desirable features required for benchmarking robustness in entity tracking or local world modeling ability. \ding{52}* indicates that the feature is partially fulfilled, while \cmark represents complete presence of the feature.}
\label{tab:datasetcmparison}
\end{table*}

\section{Effect of Pragmatic Phenomena.}
We qualitatively evaluate whether anaphora resolution is one of the factors behind low robust accuracy in \textsc{PragWorld} (manual). We specifically investigate GPT-3.5-turbo responses on the GRICE (manual) subset, and annotate each context-question as ``yes'' if they require anaphora resolution to answer, otherwise ``no''. Performance analysis shows that models correctly resolve 72 out of 129 non-anaphora cases and 47 out of 71 anaphora cases. We see that, even in cases without anaphora resolution, models still make errors, misinterpreting referents in 44.20\% of explicit scenarios. Overall, they fail to update world states consistently, leading to contradictions in 40.50\% of context-sensitive queries.

\begin{table*}[!htb]
\resizebox{\textwidth}{!}{%
\begin{tabular}{@{}lcclllllll@{}}
\toprule
 & \multicolumn{1}{l}{\textbf{\#Params}} & \multicolumn{1}{l}{\textbf{Context}} & \textbf{Robust Acc} & \textbf{Yes Acc} & \textbf{No Acc} & \textbf{Original Acc} & \textbf{Altered Acc} & \textbf{Flip Acc} & \textbf{Invariant Acc} \\ \midrule
\multicolumn{10}{c}{\textbf{\textsc{PragWorld} (manual)} } \\ \midrule
Llama-3.1-70B-Ins.                              &    70B                               &   128k                                     &  54.08 & \underline{63.54} & \textbf{97.76} & \underline{81.63} & \underline{78.11} & 76.26 & \underline{78.39}                 \\
Llama-3.3-70B-Ins.                              & 70B                                   & 128k                                      &  \textbf{61.22} & \textbf{70.76} & 93.27 & \textbf{82.65} & \textbf{80.35} & \textbf{81.29} & \textbf{79.12}

\\\midrule
Qwen2.5-32B-Ins. & 32B         & 128k              &                           54.08 & \underline{63.54} & \underline{97.31} & \underline{81.63} & 77.86 & 76.98 & 77.66             \\
Qwen2.5-72B-Ins.                           & 72B                                   & 131k                                      &   \underline{55.1} & 59.93 & \textbf{97.76} & 78.57 & 76.37 & \underline{78.42} & 74.73               \\   \bottomrule
\end{tabular}%
}
\caption{Performance of Llama-3.1-70B-Instruct, Llama-3.3-70B-Instruct, Qwen2.5-32B-Instruct, and Qwen2.5-72B-Instruct models on \datasetname{}. Best and second best performances are represented by \textit{bold}, and \textit{underline} respectively.}
\label{tab:additionalresults_globalresultstable}
\end{table*}


\bibliography{aaai2026, custom,anthology}
\end{document}